\documentclass{article}

\usepackage{iclr2026_conference,times}
\usepackage{amsmath,amssymb,amsthm}
\usepackage{booktabs}
\usepackage{graphicx}
\usepackage{subcaption}
\usepackage{xcolor}
\usepackage{array}
\usepackage[section]{placeins}
\usepackage{hyperref}
\usepackage{url}

\iclrfinalcopy


\newcommand{\Kstar}{K^\star}
\newcommand{\ncal}{n_M}
\newcommand{\nsel}{n_S}
\newcommand{\risk}{\mathcal{R}}
\newcommand{\pathvar}{\pi}
\newcommand{\schema}{s}
\newcommand{\aggfam}{\mathcal{A}}
\newcommand{\Astar}{\mathcal{A}^\star}
\newcommand{\nval}{n_V}
\newcommand{\nhuman}{n_H}
\newcommand{\fcps}{\textsc{FCPS}}

\newtheorem{theorem}{Theorem}
\newtheorem{proposition}{Proposition}
\newtheorem{corollary}{Corollary}

\title{A Finite-Calibration Regime Map for LLM Judge Panels}
\author{
Bin Zhu \quad Yi Xie \quad Yanghui Rao\thanks{Corresponding author.}\\
School of Computer Science and Engineering\\
Sun Yat-sen University, Guangzhou, China\\
\texttt{zhub35@mail2.sysu.edu.cn \quad xiey299@mail2.sysu.edu.cn}\\
\texttt{raoyangh@mail.sysu.edu.cn}
}
\date{}

\begin{document}
\maketitle

\begin{abstract}
Deploying an LLM judge panel spends human labels on fitting a calibrator, constructing
candidate judge paths, and validating which candidate to deploy. We study when
finite labels should support a low-dimensional stacker or
reliability model, and when an unrestricted joint output table is worth its
cell-count and unseen-pattern cost. We cast this as a finite-calibration regime map
and instantiate it as \emph{Finite-Calibration Panel Selection} (\fcps), a
validation selector over judge path, deployed panel size, and aggregator family
with support diagnostics. Across RewardBench, LLMBar, SummEval, and Arena100K with
a seven-judge pool, scalar/reliability aggregation has lower MSE than unrestricted
joint-table calibration in 16 of 20 real dataset--budget cells by point estimate,
while paired 95\% intervals exclude zero in 11 cells; richer
backoff/shrinkage tables narrow some gaps while preserving the finite-support
bottleneck. Controlled calibration-growth data show the opposite regime: when
labels contain a six-way interaction, the selected table grows to the
interaction-bearing prefix and its MSE falls from \(0.224\) to \(0.061\) once
unseen mass vanishes. The practical deployment question is whether the next judge's
information is estimable under the available human labels.
\end{abstract}

\section{Introduction}

Automatic judges are widely used to evaluate language model outputs
\citep{zheng2023judging,liu2023geval,zhu2023judgelm}, and judge panels are often
proposed to reduce single-judge bias \citep{verga2024juries}. The finite-label catch
is that more judges are information plus a label-allocation bill. A deployer must
spend labels to construct candidate paths or prefixes, fit the calibration map, and
validate which candidate is safe to deploy. Adding one judge may reveal a useful
interaction, or it may split the calibration table into sparse cells while
also enlarging the validation menu.

We ask when a deployer should spend the calibrator-fitting part of a finite label
budget on a flexible joint table, and when the same labels are better spent on a
lower-dimensional reliability or stacking model
\citep{wolpert1992stacked,qian2026trust}. The resulting object is an empirical
finite-calibration regime map over task structure, panel size, aggregator family, and
calibrator-fitting budget. The main experiments isolate the calibration-growth
effect by holding the selection and validation blocks fixed within each dataset and
sweeping the calibrator-fitting budget; a fixed-total-budget diagnostic then shows
how the same regime map behaves when labels are reallocated among path construction,
calibrator fitting, and validation.

For a target distribution \(P\), ordered judge path \(\pathvar\), prefix size \(K\),
output schema \(\schema\), aggregator family \(\aggfam\), selection size \(\nsel\),
calibrator-fitting size \(\ncal\), and validation size \(\nval\), let
\(\widehat f_{\pathvar,K,\aggfam}\) be the predictor fitted from the calibration
block. Its finite-sample risk is organized by the bookkeeping decomposition
\begin{equation}
  \risk(\widehat f_{\pathvar,K,\aggfam})-\risk^\star
  \approx
  \mathrm{Approx}(\pathvar,K,\aggfam)
  +
  \mathrm{Path}(\Pi,\nsel)
  +
  \mathrm{Est}(\pathvar,K,\aggfam,\ncal)
  +
  \mathrm{Sel}(\mathcal C,\nval).
\label{eq:finite-risk}
\end{equation}
Here \(\risk^\star=\inf_f \risk(f)\) is the Bayes risk, the smallest population
risk attainable by any predictor \(f\); \(\mathrm{Approx}\) is the population
approximation gap of the chosen prefix and aggregator family relative to the
Bayes rule, \(\mathrm{Path}\) is the cost
of constructing or ranking judge paths from only \(\nsel\) selection labels,
\(\mathrm{Est}\) is the calibrator-fitting error after a candidate is fixed, and
\(\mathrm{Sel}\) is the validation-selection error from choosing among the
finite candidate menu \(\mathcal C\) using \(\nval\) labels. The path-generation
menu is \(\Pi\), and \(\mathcal C\) is the finite set of fitted deployment
candidates. The selection term follows the usual finite-menu validation behavior
from Hoeffding concentration and a union bound
\citep{stone1974cross,hoeffding1963probability};
Appendix~\ref{app:proofs} makes it explicit.
Equation~\eqref{eq:finite-risk} serves as bookkeeping for label pressure across the
pipeline. The experiments therefore report observable proxies for these pressures,
such as
candidate-menu validation scale, cell pressure, unseen mass, selected prefix, and
held-out regret.

\begin{figure}[t]
  \centering
  \scriptsize
  \setlength{\tabcolsep}{3pt}
  \renewcommand{\arraystretch}{1.08}
  \begin{tabular}{>{\raggedright\arraybackslash}p{0.20\linewidth}
                  >{\raggedright\arraybackslash}p{0.34\linewidth}
                  >{\raggedright\arraybackslash}p{0.34\linewidth}}
    \toprule
     & \textbf{Low table pressure}
     & \textbf{High table pressure} \\
    \midrule
    \textbf{High interaction gain}
      & Joint-table regime: deploy the interaction-bearing prefix when validation gain and cell support improve together.
      & Use shrinkage/backoff or collect more calibration labels; deploy table structure only after validation gain clears a guard. \\
    \textbf{Low interaction gain}
      & Keep the low-dimensional stacker or reliability model unless validation shows a clear table gain.
      & Prefer scalar/reliability aggregation or a shorter prefix; table flexibility is unlikely to clear finite-cell cost. \\
    \bottomrule
  \end{tabular}
  \caption{Operational finite-calibration regime map. Table pressure is measured by
  effective support per calibrator-fitting label and unseen-pattern mass; interaction
  gain is measured by validation improvement over scalar aggregation.}
  \label{fig:operational-regime-map}
\end{figure}

The relevant decision object is
\[
  (\pathvar^\star,\Kstar,\Astar)
  \in
  \arg\min_{(\pathvar,K,\aggfam)\in\mathcal C}
  \risk(\widehat f_{\pathvar,K,\aggfam}),
\]
which selects a deployable panel/aggregator pair for the available label budget. For
the joint-table family, the first two terms reduce to the familiar squared-loss identity
\(\risk_{\mathrm{finite}}=\risk_{\mathrm{oracle}}+\risk_{\mathrm{cal}}\): the
Bayes table \(g^\star(z)=\mathbb{E}[Y\mid Z_K=z]\), where \(Z_K\) is the joint output
pattern of the first \(K\) judges, has no table-approximation error, and the finite
calibration term includes smoothing bias, cell imbalance, and fallback error on
unseen cells. Lower-dimensional stackers have smaller estimation cost but nonzero
approximation error when judge interactions matter. Thus additive targets
favor scalar stackers, while large non-additive interactions favor the joint table
after labels cover the induced cell structure.

The experiments expose this tradeoff across real and controlled regimes. On four real
benchmarks with a seven-judge pool including DeepSeek V4 Flash, validation-selected
scalar aggregation beats validation-selected joint-table prefix calibration in 16 of
20 dataset--budget cells. The deployment implication is that \fcps{} selects
low-dimensional aggregators when unrestricted joint tables lack interaction gain
relative to their finite-support diagnostics. Controlled calibration-growth regimes show the complementary
case. In an additive regime, scalar models remain better as labels grow. In a six-way
interaction regime, the selected table uses the interaction-bearing prefix and
improves sharply as unseen cells disappear. The same finite-calibration theory
explains both outcomes.

The practical output is \emph{Finite-Calibration Panel Selection} (\fcps{}): reserve
labels for path construction, calibration, and validation; generate judge paths;
train candidate aggregators for each prefix; select
\((\widehat\pathvar,\widehat K,\widehat{\aggfam})\) by validation risk; and report
effective support, exact cell pressure, and unseen rate. The contribution is the
finite-calibration regime map, deployable selector, and diagnostics that connect the
selected panel to finite-support pressure. In this view, panel complexity must be
earned by calibration and validation evidence.

\paragraph{Contribution.}
Our central contribution is a finite-calibration regime map for deploying LLM judge
panels: jointly select the judge prefix and aggregation family after accounting for
the labels spent on path construction, calibrator fitting, and validation. This has
four components.
\begin{enumerate}
  \item \textbf{Finite-calibration bottleneck.} We identify the support bottleneck
  created when added judges expand a joint output table faster than human calibration
  labels can estimate its cells.
  \item \textbf{Deployable label accounting.} We instantiate this view as \fcps{}, a
  selector over paths, prefixes, and aggregator families that separates labels for
  path construction, calibrator fitting, and validation.
  \item \textbf{Real-data regime evidence.} On four real benchmarks, scalar and
  reliability-style aggregators often win because observed interaction gains are
  small relative to table-like estimation pressure.
  \item \textbf{Interaction boundary.} Controlled and semi-real calibration-growth
  regimes show the opposite side of the map, where joint tables become useful once
  non-additive structure is present and cell support is sufficient.
\end{enumerate}

Although this paper and the companion \emph{Stopping and Routing LLM Judge
Panels} share part of the benchmark judge-output matrices and judge pool, they
address distinct deployment decisions: the companion selects conditional calls
and stopping, whereas \fcps{} selects a panel prefix and aggregation family
after candidate outputs are available.

\section{Related Work}

\paragraph{LLM-as-a-judge evaluation.}
Large language models are now widely used as automatic evaluators for open-ended
generation. MT-Bench, Chatbot Arena, G-Eval, and JudgeLM established prompted or
fine-tuned judges for pairwise, multi-turn, and NLG evaluation, while also documenting
judge biases and reliability concerns
\citep{zheng2023judging,liu2023geval,zhu2023judgelm,jain2025beyond}. Benchmarks such
as LLMBar, RewardBench, and SummEval stress instruction-following meta-evaluation,
reward/preference modeling, and summarization quality assessment
\citep{zeng2024llmbar,lambert2024rewardbench,fabbri2021summeval}; deterministic
learned metrics offer another evaluator family \citep{alam2026beyond}. We use these
outputs and benchmarks as a calibration substrate: the decision problem is how a
finite label budget should determine the deployed judge prefix and aggregator family.

\paragraph{Judge panels and finite-budget judge reliability.}
Recent work has argued that panels or ``juries'' of diverse LLM judges can outperform
a single large judge and reduce single-model bias \citep{verga2024juries}. Our results
refine this intuition: panels are useful, but the deployed panel size should still be
selected under the available calibrator-fitting and validation budget. Evidence on
correlated errors, perturbation sensitivity, and heterogeneous judge reliability
shows why panel size alone is a poor proxy for deployable information
\citep{kohli2026nine,dev2026judge,qian2026trust,yu2026heterogeneous}. Concurrent
label-efficient estimation work argues that a calibrated full panel can outperform
accuracy-based curation when the full joint signal is learnable \citep{li2026calibrate};
our finite-table diagnostics ask when that joint signal becomes too sparse for a
limited calibration set. Other multi-judge methods aggregate repeated samples or
provide abstaining finite-sample control
\citep{dadkhahi2025distribution,badshah2026scope}. Our focus is the deployable
choice of judge prefix and aggregator when the system must output calibrated
predictions for all examples.

\paragraph{Calibration and finite-sample reliability.}
Calibration has a long statistical history \citep{dawid1982calibrated} and is central
to turning model scores into reliable probabilities. Practical post-hoc calibration
methods such as Platt scaling and temperature scaling estimate a calibration map from
held-out labels \citep{platt1999probabilistic,guo2017calibration}; LLM-judge work
also studies prompting, ensembling, uncertainty, conformal intervals, surrogate
calibration, and rubric calibration
\citep{lail2026cost,radharapu2025calibrating,sheng2025analyzing,khomiakov2026noise,landesberg2025causal,feng2026noisy,hashemi2024llmrubric}.
Here the calibrated object is a joint judge-output pattern table whose support grows
with the deployed panel size, rather than a fixed scalar score or rubric
representation. The finite-table bound in Appendix~\ref{app:proofs}
makes this dependence explicit and motivates the entropy-effective support and
unseen-pattern diagnostics. Factuality-evaluation studies show that strong LLMs can
vary in judge reliability across domains \citep{fu2023reliable}, reinforcing the
need for explicit calibration and validation before deployment.

\paragraph{Model selection and ensembles.}
Validation-based model selection is standard under finite data
\citep{stone1974cross}, and stacking combines predictors to improve accuracy
\citep{wolpert1992stacked}. Judge-panel selection adds a calibration constraint:
adding a judge changes the joint-pattern alphabet estimated from human labels. The
decision is therefore whether the next judge's information gain is worth the extra
calibration complexity. Structured Dawid--Skene/Bradley--Terry and hierarchical
jury models \citep{dawid1979maximum} are richer candidate families for the same
selector: they can reduce table pressure through modeling assumptions, while still
requiring validation evidence under the available label budget.

\section{Finite-Calibration Panel Selection}

\subsection{Human-Label Accounting}

A deployment budget covers the full selection pipeline, not just the final
calibrator. If \(\nhuman\) human-labeled examples are available before deployment,
they are allocated among candidate generation, calibrator fitting, and candidate
selection:
\[
  \nhuman = \nsel + \ncal + \nval .
\]
The selection block of size \(\nsel\) constructs or ranks candidate judge paths. The
calibration block of size \(\ncal\) fits the aggregation map for each candidate. The
validation block of size \(\nval\) chooses among the resulting
\((\pathvar,K,\aggfam)\) candidates. The experiments additionally reserve a research
test block to audit the already selected decision.

This distinction matters because different terms in Eq.~\eqref{eq:finite-risk}
consume different labels. A larger candidate menu needs enough validation data to
avoid selecting a noisy winner; the formal finite-menu validation scale is given in
the next subsection. A joint table needs enough calibration data to cover its induced
cells, with pressure measured by
\(\sum_z p_z/N_z\) and unseen mass. Thus, under a fixed \(\nhuman\), a deployer can
either reserve more labels for validation, shrink the candidate menu, or bias the
selector toward simpler aggregators through a one-standard-error or guarded rule.

The main experiments fix \(\nsel\), \(\nval\), and the held-out test size for each
dataset, then sweep \(\ncal\). This calibration-growth design isolates how additional
calibrator-fitting labels change cell pressure and the scalar-versus-table regime.
The separate fixed-\(\nhuman\) allocation diagnostic complements this view by moving
labels among \(\nsel\), \(\ncal\), and \(\nval\) while keeping the same deployment
accounting.

\subsection{Decision Object}

A deployment decision consists of a judge path \(\pathvar\), a prefix size \(K\), and
an aggregator family \(\aggfam\). The family \(\aggfam\) specifies how labeled
calibration examples are used to map the first \(K\) judge outputs to a calibrated
prediction. In this paper, the main nonparametric family is the joint table in
Eq.~\eqref{eq:smooth-estimator}; the main low-dimensional alternatives are mean
aggregation with isotonic calibration, ridge stacking with isotonic calibration, and
logistic stacking, following standard post-hoc calibration and stacking ideas
\citep{platt1999probabilistic,guo2017calibration,wolpert1992stacked}.

The finite-calibration decision is
\begin{equation}
  (\widehat{\pathvar},\widehat K,\widehat{\aggfam})
  \in
  \arg\min_{\pathvar\in\Pi,\,K\in\{1,\ldots,m\},\,\aggfam\in\mathfrak A}
  \widehat{\risk}_{\mathrm{val}}(\widehat f_{\pathvar,K,\aggfam}).
  \label{eq:general-selection}
\end{equation}
The original path/prefix selector is the special case where
\(\mathfrak A=\{\text{joint table}\}\). The scalar aggregation baselines and the
controlled regime experiment use the same validation principle while varying
\(\aggfam\). This \fcps{} procedure is a finite-candidate validation selector
\citep{stone1974cross}. Its contribution is the finite-calibration candidate space
over paths, prefix sizes, and aggregation families together with the diagnostics
used to interpret the selected regime.

The formal validation bound in Appendix~\ref{app:proofs} is the standard finite-menu
concentration result: for bounded losses, selecting over \(|\mathcal C|\) fitted
candidates satisfies
\[
  \risk(\widehat f_{\widehat c})
  \le
  \min_{c\in\mathcal C}\risk(\widehat f_c)
  +
  O\!\left(\sqrt{\frac{\log|\mathcal C|}{\nval}}\right).
\]
Thus larger candidate menus are statistically affordable only when the validation
block has enough resolution, or when the menu is guarded toward simpler choices.
In the main real-data comparisons, the joint-table menu has \(3\times 7=21\)
candidates and the scalar-family menu has \(6\times 7=42\); Appendix
Table~\ref{tab:candidate-menu-validation} reports the corresponding
\(\sqrt{\log|\mathcal C|/\nval}\) scales.

\subsection{Judge Paths and Prefixes}

Let \(\mathcal{J}=\{j_1,\ldots,j_m\}\) be a pool of judges. A path rule produces an
ordering \(\pathvar\) of this pool. For each prefix size \(K\), the deployed panel is
the first \(K\) judges along the path. The path rule therefore determines which joint
pattern table is exposed as \(K\) grows.

We consider three path rules. The \emph{information-first path} orders judges by
single-judge oracle performance on the selection block and is the strong path
baseline. The \emph{random path} is a null baseline. The
\emph{complexity-penalized path} greedily trades selection-block oracle gain against
the increase in entropy-effective support, normalized by the planned
calibrator-fitting budget \(\ncal\). Path rules only propose candidates from the
selection block; calibration and validation fit and select the deployed predictor.
The appendix gives the full path heuristic and an additional calibration-frontier
diagnostic.

\subsection{Calibration Pressure and Aggregator Complexity}

The calibration-complexity term in Eq.~\eqref{eq:finite-risk} is most visible for
joint output tables. For a fixed \((P,K,\pathvar,\schema)\), let \(Z_K\) be the joint
output pattern of the first \(K\) judges. A table calibrator estimates
\(g^\star(z)=\mathbb{E}[Y\mid Z_K=z]\) from \(\ncal\) calibration labels. If
\(N_z\) is the calibration count in cell \(z\), the formal bound in
Appendix~\ref{app:proofs} shows that the table pays a weighted small-cell term plus
unseen mass. For the calibration block \(D_M\), the operational pressure is
\begin{equation}
  V_K(D_M)
  =
  \sum_{z:N_z>0} p_z N_z^{-1}
  +
  P(Z_K\notin\widehat{\mathcal Z}_K),
  \label{eq:cell-pressure}
\end{equation}
up to logarithmic, smoothing, and bounded-loss constants. In the experiments we
report a retrospective held-out plug-in audit
\(\widehat V^{\mathrm{cell}}_K=\sum_{z:N_z>0}\widehat p_{\mathrm{test}}(z)/N_z+
\widehat u_K\), where \(\widehat u_K\) is the held-out unseen-pattern rate; the audit
uses calibration cell counts and held-out pattern frequencies. We
also report \(\widetilde V_K=H_K/\ncal+\widehat u_K\), where
\(H_K=\exp(-\sum_z \widehat p_z\log \widehat p_z)\) is the entropy-effective
occupied support. In deployment, the same inverse-count term can be estimated from
unlabeled target traffic after the candidate panel has been run, by replacing
\(p_z\) with target pattern frequencies \(\widehat q(z)\); this requires judge
outputs but no additional human labels. When such target traffic is unavailable,
\(\widetilde V_K\) is the cold-start proxy rather than an exact deployment
criterion.

This pressure creates an approximation--estimation boundary. Low-dimensional
aggregators have much smaller estimation cost but can miss interactions; joint tables
represent arbitrary categorical interactions but need enough calibration labels to
cover the induced support. The formal table and stacking statements in
Appendix~\ref{app:proofs} make the scale explicit. In the approximately balanced
case with \(M_K\) reachable output patterns, the joint table has estimation pressure
of order \(E_{\mathrm{tab}}(K,\ncal)\asymp M_K\log M_K/\ncal\), up to constants,
smoothing, and logarithmic confidence factors, and unrestricted tables have a
matching \(M_K/\ncal\)-type lower bound. A bounded linear stacker over \(K\) scalar
judge features instead has generic bounded-loss scale
\[
  E_{\mathrm{scal}}(K,\ncal)=O\!\left(\sqrt{K/\ncal}\right),
  \qquad
  E_{\mathrm{tab}}(K,\ncal)\asymp M_K\log M_K/\ncal .
\]
Thus, if \(\Delta_K(\aggfam)\) is the approximation gap of a scalar family relative
to the unrestricted table Bayes rule, a table is justified only when
\(\Delta_K(\aggfam)\gtrsim
E_{\mathrm{tab}}(K,\ncal)-E_{\mathrm{scal}}(K,\aggfam,\ncal)\).
We use this diagnostic inequality to organize the regime map behind the validation
menu and diagnostics, summarized operationally in
Figure~\ref{fig:operational-regime-map}. It has three direct deployment consequences.
First, when validation labels are scarce, the candidate menu should be small: adding
many paths, prefixes, and aggregator families only helps if \(\nval\) can resolve
their validation risks. Second, when calibration labels are scarce relative to the
selected table support, the finite-label response is to prefer a lower-dimensional
family, a shorter prefix, or a shrinkage/backoff table until unseen mass and
small-cell pressure are low. Third, a joint table is
appropriate when it clears both tests at once: it shows validation improvement over
scalar competitors and its selected cells are sufficiently observed. The same prefix
can therefore be reasonable under a scalar stacker and unreasonable under a joint
table, because the two families spend calibration labels differently.

These consequences give the practical complexity ladder used below: mean/vote plus
calibration, one-coin reliability, ridge/logistic stacking, pairwise-feature
stacking, and finally the joint table. The formal result covers the table endpoint
and bounded linear/additive stackers; one-coin reliability and isotonic
post-calibration are practical comparators, and pairwise features follow the same
\(O(K^2)\) parametric intuition. The theory therefore organizes the candidate
families and diagnostics around their calibration-label demand. Additive or
redundant judge signals should favor stacking under finite labels, whereas important
output-pattern interactions justify a joint table only after the calibrator-fitting
budget can estimate the relevant cells.

\subsection{\fcps{} Protocol}

Given a deployable labeled budget split into \((\nsel,\ncal,\nval)\), \fcps{} uses
the selection block to generate paths, the calibration block to fit aggregators, and
the validation block to choose the deployed candidate. For each candidate path rule
and prefix size \(K\), we fit a calibration map on the calibration block and evaluate
finite risk on the validation block. For the joint-table experiments, the selected
prefix is
\begin{equation}
  (\widehat{\pathvar}, \widehat{K})
  \in
  \arg\min_{\pathvar \in \Pi,\, K \in \{1,\ldots,m\}}
  \widehat{\risk}_{\mathrm{val}}(\widehat f_{\pathvar,K,\mathrm{tab}}).
  \label{eq:selection}
\end{equation}
This is Eq.~\eqref{eq:general-selection} restricted to
\(\mathfrak A=\{\text{joint table}\}\). In deployment, after validation selection,
the chosen predictor is frozen or refit on the permitted labeled data according to
the deployment policy. In the experiments, a separate research test block evaluates
the already selected decision and records diagnostic quantities such as unseen
pattern rate.

This framing separates \fcps{} from any single path generator or aggregator. In the
real-data experiments, joint-table path/prefix selection and scalar-family selection
use the same calibration/validation/test separation, so
Tables~\ref{tab:baseline-envelope}--\ref{tab:regime-comparison} compare families under
a common validation-selection protocol. Adding a judge also changes the calibration
problem itself: it expands the joint pattern support that must be estimated before a
table or table-like residual can be safely deployed.

In the present implementation, the calibration map is a finite pattern-table
estimator. Each observed joint judge-output pattern is mapped to its empirical
calibration target on the calibration block, smoothed toward the calibration marginal
with \(\alpha=0.5\):
\begin{equation}
  \widehat{g}(z)
  =
  \frac{\sum_{i:Z_i=z} y_i + \alpha \bar{y}_{\mathrm{cal}}}{N_z+\alpha}.
  \label{eq:smooth-estimator}
\end{equation}
Patterns not observed during calibration are assigned the fallback prediction
\(\bar{y}_{\mathrm{cal}}\). We report the unseen pattern rate because this fallback is
exactly where finite calibration pressure becomes visible. Risk is the mean squared
calibration error, \(\frac{1}{|D|}\sum_{i\in D}(y_i-\widehat{g}(Z_i))^2\), on the
validation or test block. All path construction and prefix selection use only the
selection/calibration/validation blocks; the test block is held out for final
evaluation and diagnostics.
With its global fallback, this joint table is a nonparametric endpoint; hierarchical
backoff, Bayesian smoothing, and residual shrinkage are middle families between
scalar stackers and a fully unrestricted table \citep{dawid1982calibrated,wolpert1992stacked}.

\paragraph{Algorithmic protocol.}
Operationally, \fcps{} partitions labels into selection, calibration, and validation
blocks; generates paths on the selection block; fits each
\((\pathvar,K,\aggfam)\) candidate on calibration data; selects by validation risk
using Eq.~\eqref{eq:general-selection}; and deploys the selected candidate when its
validation evidence and complexity diagnostics support the added complexity. In
evaluation mode, a separate test block audits the already selected decision.

\paragraph{Practical allocation rule.}
For moderate label budgets, a simple default is to allocate roughly
\(\nsel:\ncal:\nval=1:2:1\), so the calibrator receives the largest block while the
validation selector still has enough resolution to compare candidates. If the
candidate menu is large, either increase \(\nval\) or shrink the menu before
validation. If joint-table interactions are strongly suspected, increase \(\ncal\)
and require low unseen mass before deployment. In cold-start regimes with only a few
dozen labels, the guarded variant in Appendix~\ref{app:extended-results} starts from
a task-type prior and adapts when validation evidence clears a
one-standard-error-style guard.

\paragraph{Hybrid shrinkage extension.}
The same pressure also motivates count-shrunk residual tables: add a validation-tuned
residual correction to a scalar predictor only in well-observed cells, and fall back
to the scalar model in sparse cells. Appendix~\ref{app:extended-results} reports this
diagnostic alongside the pre-specified table/scalar menu, where the strongest scalar
selector remains the most stable real-data comparator.

\subsection{Complexity Diagnostics}

The calibration-complexity term is operationalized through effective support and
unseen pattern diagnostics. For a prefix of size \(K\), \(H_K\) denotes the
entropy-effective occupied support from the natural-log definition above, and \(Q_K\)
denotes the nominal pattern alphabet size. We use \(H_K/\ncal\) as a practical proxy
for finite-budget pattern pressure and report
\(\widetilde V_K=H_K/\ncal+\widehat u_K\) when we want a single observable diagnostic
for the table pressure in Eq.~\eqref{eq:cell-pressure}. We also measure unseen
pattern rate relative to the calibration support. In research evaluation,
Table~\ref{tab:exact-cell-pressure} computes the closest audit using the held-out
test block after selection is fixed. In deployment, the analogous audit should be
computed from unlabeled target traffic pattern frequencies; if no such traffic is
available, \(H_K/\ncal+\widehat u_K\) should be treated as a rough proxy for
sparse-cell pressure. These quantities serve as observable audits of regimes in which
the literal table-bound quantities, especially small cell counts and unseen mass, are
likely to matter.

\section{Experiments}

We evaluate \fcps{} on RewardBench, LLMBar, SummEval, and
Arena100K \citep{lambert2024rewardbench,zeng2024llmbar,fabbri2021summeval,tang2025arenaexplorer,chiang2024chatbotarena}
with a fixed seven-judge pool: Qwen2.5-7B, Llama-3.1-8B, Mistral-7B,
Prometheus-7B, Gemma-3-12B, Selene-8B, and DeepSeek V4 Flash
\citep{qwen2024qwen25,grattafiori2024llama3,jiang2023mistral,kim2024prometheus2,gemmateam2025gemma3,alexandru2025selene,deepseekai2026deepseekv4}. Pairwise datasets use
categorical direct-preference judge outputs; SummEval uses discrete scalar rubric
bins. Arena100K is a filtered English non-tie LMArena-100K subset for calibrated
preference prediction. All results use 30 grouped splits by source
identifier. Each split contains a path-selection block, a calibration-fitting block of size
\(\ncal\), a validation block, and a held-out test block; candidate paths are built
on selection data, aggregators are fit on calibration data, \((K,\aggfam)\) is
selected on validation data, and test labels are reserved for final evaluation and
diagnostics. The budget \(\ncal\) in Table~\ref{tab:protocol} is the
calibrator-fitting block within the full label allocation. The main real-data tables
are calibration-growth curves under fixed selection and validation capacity.
The block sizes in Table~\ref{tab:protocol} are chosen to isolate calibration
growth rather than optimize a benchmark-specific split: selection and validation
are fixed while \(\ncal\) varies, and the held-out test block is used only as a
lower-noise audit of the selected candidate. Absolute sizes are scaled to each
dataset so grouped splits fit without separating source-linked examples.
Table~\ref{tab:fixed-budget-allocation} separately checks that the qualitative
conclusion is not tied to this single allocation.
Table~\ref{tab:fixed-budget-allocation}
adds a fixed-\(\nhuman\) allocation check that reassigns labels among selection,
calibration, and validation while keeping the research test block outside the
deployable budget.
Appendix~\ref{app:external-binary-validation} uses the same seven-judge protocol as
an out-of-domain check on three binary targets: safety, math correctness, and code
correctness \citep{chao2024jailbreakbench,cobbe2021training,austin2021program}.

\begin{table}[!htbp]
\centering
\small
\begin{tabular}{lccc}
\toprule
Dataset & calibrator-fitting budgets \(\ncal\) & select/validation/test & schema \\
\midrule
RewardBench & 50, 100, 200, 400, 800 & 300 / 300 / 600 & pairwise \\
LLMBar & 20, 50, 100, 200, 300 & 100 / 100 / 200 & pairwise \\
SummEval & 32, 64, 128, 256, 512, 800 & 192 / 192 / 384 & scalar \\
Arena100K & 50, 100, 200, 400 & 200 / 200 / 400 & pairwise \\
\bottomrule
\end{tabular}
\caption{Experimental protocol. Each row uses 30 grouped splits, smoothing
\(\alpha=0.5\), and prefix sizes \(K\in\{1,\ldots,7\}\). The block \(\ncal\) is
the calibrator-fitting budget swept in the main experiments; the selection,
validation, and test block sizes are fixed per dataset.}
\label{tab:protocol}
\end{table}

The joint-table calibrator maps each observed judge-output pattern to a smoothed cell
mean, \((N_z\bar Y_z+\alpha\bar y_{\mathrm{cal}})/(N_z+\alpha)\), and backs off to
the calibration marginal on unseen patterns. Scalar comparators include mean/vote
aggregation with isotonic calibration, one-coin reliability aggregation, ridge
stacking with isotonic calibration, logistic stacking for pairwise targets, and
pairwise-feature variants \citep{platt1999probabilistic,guo2017calibration,wolpert1992stacked}.
The one-coin comparator is supervised rather than a full Dawid--Skene EM fit.
After binarizing labels and scalarized verdicts at \(1/2\), each judge receives a
single symmetric log-odds weight from its Laplace-smoothed calibration accuracy
(clipped to \([0.05,0.95]\)); the weighted vote is combined with a
Laplace-smoothed class prior and then isotonic-calibrated.
For scalar aggregators on pairwise tasks, verdicts are encoded as \(A/a=1\),
\(B/b=0\), and tie or parse-error outcomes as \(1/2\); the joint table keeps the
categorical outputs as pattern values. This encoding makes simple mean/vote
aggregation well behaved on RewardBench and Arena100K, while LLMBar still requires
learned reliability or stacking weights. Appendix~\ref{app:extended-results} adds
candidate-menu accounting, a representative unsupervised heterogeneous-jury
baseline, negative log-likelihood (NLL) selection, tie/error encoding ablations,
fixed-\(\nhuman\) allocation checks, judge-pool perturbation, fallback/smoothing
sensitivity, deterministic prompt perturbation, and hierarchical-backoff tables. Each reported deployable comparison is an \fcps{}
restriction to a candidate menu: joint-table only, scalar-family only, or the union
of path and aggregator choices. Mean squared error (MSE) is the validation objective
tied to the theory; on pairwise tasks it is the Brier score for probabilistic binary
prediction, a bounded strictly proper score. NLL, accuracy, Pearson, and Spearman are
robustness checks reported in the appendix.

\FloatBarrier

\paragraph{Reading the diagnostics.}
Table~\ref{tab:exact-cell-pressure} audits the estimation bill paid by the selected
joint table: when \(\sum_z \widehat p_{\mathrm{test}}(z)/N_z\) or \(\widehat u\) is
large, the selected table requires more calibration support before its flexibility is
useful. Appendix Table~\ref{tab:baseline-envelope} gives the matching path/envelope
comparison: selected prefixes usually improve over the all-seven table, while the
non-deployable test oracle remains lower, identifying validation as a statistical
bottleneck. At the largest budgets,
path/prefix validation improves over the all-seven table
(\(0.025/0.028\), \(0.203/0.211\), \(0.045/0.059\), \(0.233/0.245\)),
while the test-oracle envelope remains lower
(\(0.022,0.194,0.043,0.227\)).

\section{Results}

The empirical results make the finite-calibration map concrete under fixed
selection and validation capacity. First, within the joint-table restriction of
\fcps, risk over \(K\) often flattens or turns upward, and selected prefixes are
usually smaller than the full seven-judge pool
(Appendix Figures~\ref{fig:risk-k}--\ref{fig:validation-risk}). Table~\ref{tab:exact-cell-pressure}
tracks cell pressure: it falls as the calibrator-fitting block grows, while the
entropy proxy can understate literal small-cell pressure. Controlled residualized
checks support the proxy: standardized \(H_K/\ncal\) and unseen-rate coefficients for
test risk are \(0.178\) and \(0.199\), both with standard error \(0.009\)
(Appendix Table~\ref{tab:controlled-complexity}).

\begin{table}[!htbp]
\centering
\scriptsize
\begin{tabular}{lrrrrrrrr}
\toprule
Dataset & $n_M$ & $\widehat K$ & $\sum\hat p_z/N_z$ & $\hat u$ & $\widehat V^{cell}$ & proxy $\widetilde V$ & table & scalar \\
\midrule
RewardBench & 50 & 2.57 & 0.07510 & 0.0334 & 0.10855 & 0.0885 & 0.0259 & 0.0238 \\
RewardBench & 800 & 3.63 & 0.02282 & 0.0077 & 0.03049 & 0.0127 & 0.0251 & 0.0225 \\
LLMBar & 20 & 2.37 & 0.21937 & 0.1020 & 0.32137 & 0.3266 & 0.2257 & 0.2285 \\
LLMBar & 300 & 4.13 & 0.09100 & 0.0292 & 0.12017 & 0.0749 & 0.2031 & 0.1857 \\
SummEval & 32 & 1.47 & 0.24489 & 0.0746 & 0.31946 & 0.2565 & 0.0559 & 0.0511 \\
SummEval & 800 & 2.47 & 0.06106 & 0.0169 & 0.07799 & 0.0474 & 0.0454 & 0.0435 \\
Arena100K & 50 & 2.43 & 0.10241 & 0.0405 & 0.14291 & 0.1129 & 0.2395 & 0.2383 \\
Arena100K & 400 & 2.73 & 0.03033 & 0.0067 & 0.03700 & 0.0195 & 0.2331 & 0.2324 \\
\bottomrule
\end{tabular}
\caption{Post-selection held-out audit of cell pressure for the validation-selected joint-table candidate. The column $\widehat V^{cell}=\sum_{z:N_z>0}\widehat p_{\mathrm{test}}(z)/N_z+\widehat u$ directly instantiates the two terms in Eq.~\eqref{eq:cell-pressure} using calibration cell counts and held-out test pattern frequencies. The selector itself uses validation risk; the proxy $\widetilde V=H_{\widehat K}/n_M+\widehat u$ is shown for comparison, followed by the validation-selected joint-table and scalar-family test risks.}
\label{tab:exact-cell-pressure}
\end{table}

\begin{figure}[t]
  \centering
  \begin{subfigure}{0.49\linewidth}
    \centering
    \includegraphics[width=0.84\linewidth]{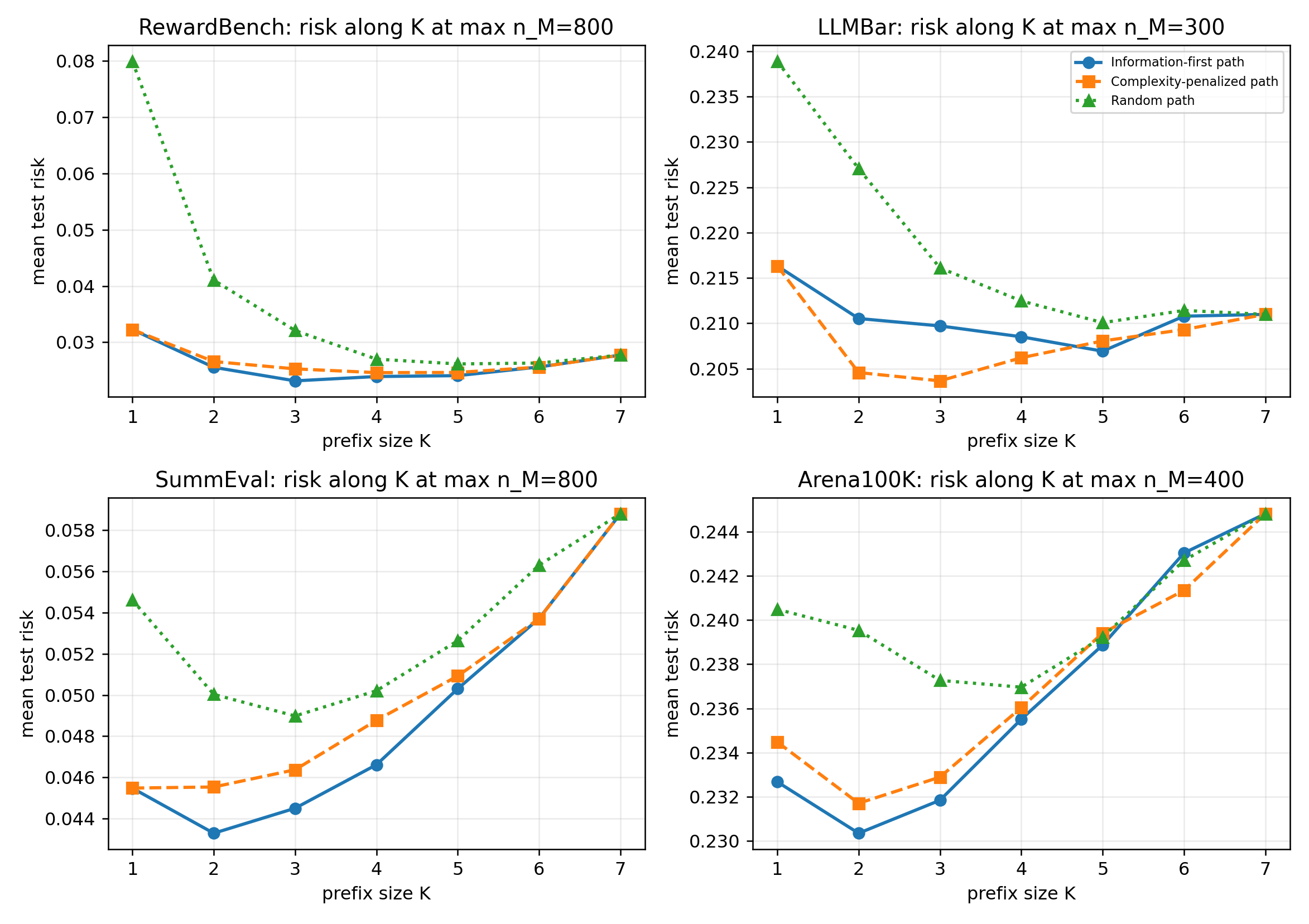}
    \caption{Risk over prefix size.}
  \end{subfigure}
  \hfill
  \begin{subfigure}{0.49\linewidth}
    \centering
    \includegraphics[width=0.84\linewidth]{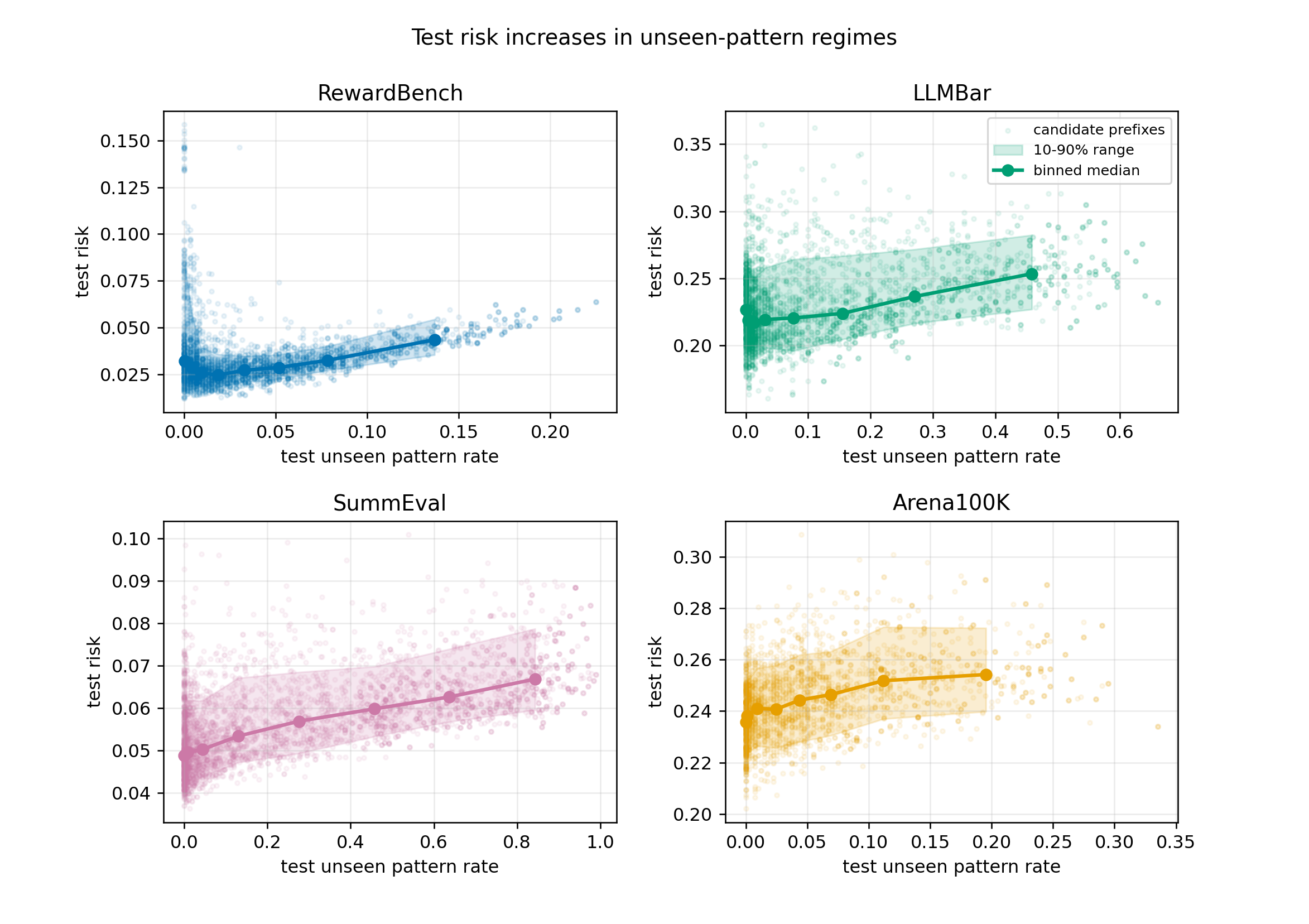}
    \caption{Risk versus unseen-pattern rate.}
  \end{subfigure}
  \caption{Main finite-calibration regime diagnostics. Left: joint-table risk often
  flattens or worsens as more judges increase pattern support. Right: held-out risk
  rises with unseen-pattern pressure, so table-like expansion must clear both
  validation and support diagnostics.}
  \label{fig:main-regime-diagnostics}
\end{figure}

Second, current real benchmarks usually lie on the low-interaction,
finite-support side of the map. Scalar aggregation has lower mean test MSE in
16 of 20 dataset--budget comparisons against joint-table prefix calibration, with
paired 95\% intervals excluding zero in 11 cells (Appendix
Table~\ref{tab:regime-comparison}); margins are largest on RewardBench and LLMBar
and small on SummEval and Arena100K. Table~\ref{tab:main-finite-label-checks}
gives the mechanism audit at the largest budgets: pairwise-feature stackers add
little over additive stackers, pairwise features are selected in only
\(0.10\)--\(0.33\) of scalar splits, and the non-deployable joint-table oracle is
not lower than the scalar oracle. Thus the table usually pays support and
validation cost before there is enough interaction gain to learn. Richer
table-like menus--hierarchical backoff and residual shrinkage--narrow some gaps
and are nominally better in small-margin cells, but RewardBench and LLMBar remain
scalar-favored under the same split protocol (Appendix
Tables~\ref{tab:table-like-family-comparison},
\ref{tab:oracle-interaction-gap-full}, and
\ref{tab:backoff-split-sensitivity}).

\begin{table}[!htbp]
\centering
\scriptsize
\resizebox{\linewidth}{!}{%
\begin{tabular}{lrrrrrr}
\toprule
Dataset & $\widetilde V$ & pairwise gain & pairwise sel. & oracle $\Delta$ & finite $\Delta$ & residual $\Delta$ \\
\midrule
RewardBench & 0.0127 & -0.0000 & 0.30 & +0.0021 & +0.0025 & -0.0004 \\
LLMBar & 0.0749 & -0.0024 & 0.30 & +0.0145 & +0.0205 & +0.0007 \\
SummEval & 0.0474 & -0.0004 & 0.33 & -0.0000 & +0.0019 & +0.0002 \\
Arena100K & 0.0195 & -0.0013 & 0.10 & +0.0014 & +0.0010 & +0.0003 \\
\bottomrule
\end{tabular}%
}
\caption{Mechanism audit for why scalar aggregation usually wins on the real-data regime at each dataset's largest calibrator-fitting budget. $\widetilde V=H_{\widehat K}/n_M+\widehat u$ is table pressure for the validation-selected table; pairwise gain is additive-stack test risk minus pairwise-feature test risk for the all-seven scalar stacker, so positive values favor interaction features; pairwise sel. is the share of validation-selected scalar aggregators that use pairwise features. Oracle and finite $\Delta$ are joint-table risk minus scalar risk, so positive values favor scalar aggregation. Residual $\Delta$ is residual-shrinkage minus scalar risk. Full oracle gaps for all dataset--budget cells are in Appendix Table~\ref{tab:oracle-interaction-gap-full}.}
\label{tab:main-finite-label-checks}
\end{table}

Sensitivity checks support the same regime reading without making it a single
downstream convention. Pool perturbation from the six-judge Selene pool to the
seven-judge DeepSeek pool changes risk levels, especially on LLMBar and
Arena100K, but selected prefixes remain below the full pool (Appendix
Table~\ref{tab:pool-artifact-guardrail}). A deterministic prompt perturbation on
four general local LLMBar judges changes about one third of individual verdicts
while retaining sub-full selected prefixes (Appendix
Table~\ref{tab:prompt-perturbation}). Fallback/smoothing, tie/error encoding,
NLL/metric checks, and fixed-\(\nhuman\) reallocations give the same qualitative
scalar/table reading (Appendix Tables~\ref{tab:table-sensitivity},
\ref{tab:tie-parse-fairness}, \ref{tab:nll-selection-ablation},
\ref{tab:metric-robustness}, and \ref{tab:fixed-budget-allocation}). Together,
these diagnostics place the real-data benchmarks in a regime where table
complexity should be spent only when validation gain and cell support improve
together.

Third, controlled and semi-real interaction stress tests show the opposite regime.
At \(\ncal=1024\), the scalar-minus-table gaps in the synthetic/semi-real
calibration-growth checks are \(-0.002/-0.002\) for additive targets,
\(+0.018/+0.006\) for mixed targets, and \(+0.053/+0.015\) for parity targets
(Appendix Tables~\ref{tab:calibration-growth-new-dataset} and
\ref{tab:semireal-calibration-growth}). Thus additive structure remains scalar-
favored, while mixed and high-order parity structure becomes table-favored once the
interaction is both present and estimable. In the six-way synthetic parity dataset,
the selected table uses \(K=6\), unseen mass falls from \(0.768\) at \(\ncal=16\) to
zero by \(\ncal=512\), and test MSE drops from \(0.224\) to \(0.061\).
Appendix~\ref{app:extended-results} also reports robustness checks and cold-start
deployment, where guarded selection starts from a task-type prior and adapts when
few-label validation supports the change. On external binary safety, math, and code tasks, the
same seven-judge prefix protocol selects intermediate average prefixes
\(K=4.50,2.17,2.63\), while \(K=1\to7\) unseen rate grows to
\(0.210,0.033,0.053\)
(Appendix Table~\ref{tab:external-binary-validation-results}).

\paragraph{Practical deployment reading.}
These results suggest a conservative workflow for the reported menus and split
ratios: start with a strong single judge, mean/reliability aggregation, and one or
two regularized stackers; add pairwise features or a joint table only when validation
beats the scalar menu and cell pressure is low. If a table wins but pressure is
high, the next labels should usually go to calibration or shrinkage/backoff; if many
candidates overlap, the bottleneck is \(\nval\).

\section{Conclusion}

Judge-panel design is a finite-label accounting problem: a stronger judge or larger
pool can add information, but also consumes candidate-generation, calibration, and
validation capacity. The deployment question is whether the next judge exposes
structure that the available calibrator-fitting labels can learn and the validation
labels can select. On current real benchmarks, scalar and reliability aggregators
often capture most useful signal under the reported protocol, although many margins
are small; in controlled and semi-real interaction regimes, joint tables become
valuable once unseen mass and small-cell pressure fall. \fcps{} validates over a
compact menu and audits whether the selected complexity is supported by finite
labels, so panel expansion enters deployment as a validated complexity choice.

\section*{Acknowledgements}
This work was supported by the National Natural Science Foundation of China
(62372483).

\appendix

\section{Limitations}

The real-data study targets calibration-scale deployment decisions for a fixed
seven-judge pool. The main experiments sweep the calibrator-fitting block \(\ncal\)
while holding the selection and validation blocks fixed per dataset; the
fixed-\(\nhuman\) diagnostic then varies a compact set of allocation ratios. A fuller
deployment study would jointly vary \((\nsel,\ncal,\nval)\), candidate-menu size,
and total label budget. The present runs also use deterministic judge outputs, while
larger deployments can add domain-specific or frontier judges, increase pool
diversity, and sample each judge stochastically. Many benchmarks in the current suite
appear additive or redundant, and some margins are small. The theory-proxy diagnostic
matches 16 of 20 dataset--budget cells, while split-level discrimination remains
modest. The appendix adds a representative Dawid--Skene/BT-style jury baseline,
hierarchical backoff, and residual-shrinkage diagnostics; full BTD variants,
conformal abstention, dynamic jury selection, and more expressive higher-order
stackers can be added as richer \fcps{} candidate families. Heterogeneous-jury
methods that infer latent item quality or unsupervised judge reliability from
comparison graphs complement the fixed judge-output calibration studied here. Larger,
more heterogeneous judge pools are needed to test how the regime tradeoff scales.

\section{Proofs}
\label{app:proofs}

\begin{theorem}[Finite validation selection]
\label{thm:validation-selection}
Let \(\mathcal C\) be a finite set of candidate deployment choices
\((\pathvar,K,\aggfam)\). For each \(c\in\mathcal C\), let \(\widehat f_c\) be the
predictor produced from the selection and calibration blocks, and let
\(\widehat c=\arg\min_{c\in\mathcal C}\widehat{\risk}_{\mathrm{val}}(\widehat f_c)\)
be the validation selector. If the loss is bounded in \([0,1]\), then conditional on
the fitted candidate predictors, with probability at least \(1-\delta\) over the
validation block,
\begin{equation}
  \risk(\widehat f_{\widehat c})
  \le
  \min_{c\in\mathcal C}\risk(\widehat f_c)
  +
  2\sqrt{\frac{\log(2|\mathcal C|/\delta)}{2\nval}}.
  \label{eq:validation-oracle}
\end{equation}
\end{theorem}

\begin{proposition}[Joint-table estimation with unseen mass]
\label{prop:table}
Fix a prefix \(K\), path \(\pathvar\), and schema \(\schema\). Let
\(\widehat{\mathcal Z}_K=\{z:N_z>0\}\) be the occupied calibration support,
\(S_K=|\widehat{\mathcal Z}_K|\), \(p_z=P(Z_K=z)\), and
\(u_K=P(Z_K\notin\widehat{\mathcal Z}_K)\) be the test-time unseen mass. Suppose
\(Y\in[0,1]\), predictions are clipped to \([0,1]\), and the table uses the smoothed
estimate
\[
  \widetilde g(z)=\frac{N_z\bar Y_z+\alpha\bar y_{\mathrm{cal}}}{N_z+\alpha}
\]
on occupied cells and the fallback \(\bar y_{\mathrm{cal}}\) on unseen cells. With
probability at least \(1-\delta\) conditional on the calibration inputs and the
resulting occupied support, uniformly over occupied cells,
\[
  \bigl|\bar Y_z-g^\star(z)\bigr|
  \le
  \sqrt{\frac{\log(2S_K/\delta)}{2N_z}}.
\]
Consequently, the squared-loss excess risk of the deployed table relative to the
Bayes table satisfies
\begin{equation}
  \risk(\widetilde g)-\risk(g^\star)
  \le
  2\sum_{z\in\widehat{\mathcal Z}_K}
  p_z\frac{\log(2S_K/\delta)}{2N_z}
  +
  2\sum_{z\in\widehat{\mathcal Z}_K}
  p_z\left(\frac{\alpha}{N_z+\alpha}\right)^2
  +
  u_K .
  \label{eq:table-integrated-bound}
\end{equation}
\end{proposition}

\begin{proposition}[Balanced-support table scale]
\label{prop:balanced-table}
Fix \(K,\pathvar,\schema\), and suppose the population support of \(Z_K\) has
size \(M_K\). Assume approximate balance: for a constant \(c\ge 1\),
\[
  \frac{1}{cM_K}\le p_z\le \frac{c}{M_K}
  \qquad\text{for every support cell }z .
\]
If \(\ncal \ge C c M_K\log(M_K/\delta)\), then with probability at least
\(1-\delta\), all population cells are observed and the joint-table excess risk in
Proposition~\ref{prop:table} is at most
\begin{equation}
  \risk(\widetilde g)-\risk(g^\star)
  \le
  C' c^2\,\frac{M_K\log(M_K/\delta)}{\ncal}
  + O\!\left(\frac{\alpha^2 M_K}{\ncal^2}\right).
  \label{eq:balanced-table-rate}
\end{equation}
Conversely, for Bernoulli targets with conditional variance bounded below by
\(\sigma^2>0\) on each cell, estimating an unrestricted table over \(M_K\) cells has
minimax expected squared excess risk at least
\[
  \Omega\!\left(\frac{\sigma^2 M_K}{\ncal}\right)
\]
in this balanced-support regime.
\end{proposition}

\begin{corollary}[Approximation--estimation boundary]
\label{cor:regime-boundary}
For a scalar aggregator family \(\aggfam\), define its approximation gap relative to
the unrestricted table Bayes rule as
\[
  \Delta_K(\aggfam)
  =
  \inf_{f\in\aggfam}\risk(f)-\risk(g^\star_K),
\]
where \(g^\star_K(z)=\mathbb E[Y\mid Z_K=z]\). Let \(E_{\mathrm{tab}}(K,\ncal)\)
denote the table estimation pressure, e.g. the right-hand side of
Eq.~\eqref{eq:balanced-table-rate}, and let
\(E_{\mathrm{scal}}(K,\aggfam,\ncal)\) denote the scalar-family estimation pressure,
e.g. Eq.~\eqref{eq:parametric-bound}. Up to constants and validation-selection
penalties, the joint table is preferable when
\begin{equation}
  \Delta_K(\aggfam)
  \gtrsim
  E_{\mathrm{tab}}(K,\ncal)-E_{\mathrm{scal}}(K,\aggfam,\ncal),
  \label{eq:table-win-condition}
\end{equation}
whereas the scalar family is preferable when the inequality is reversed.
\end{corollary}

\begin{proposition}[Parametric stacking estimation scale]
\label{prop:stacking}
Let \(x\in[0,1]^K\) be the vector of scalarized judge outputs for a \(K\)-judge
prefix, and let \(\mathcal F_{K,B}=\{x\mapsto \phi(w^\top x+b):\|w\|_2\le B,\,
|b|\le B\}\), where \(\phi\) is a 1-Lipschitz clipping or sigmoid map. For any
\(L\)-Lipschitz loss bounded in \([0,1]\), empirical risk minimization over
\(\mathcal F_{K,B}\) satisfies, with probability at least \(1-\delta\),
\begin{equation}
  \risk(\widehat f)
  \le
  \inf_{f\in\mathcal F_{K,B}}\risk(f)
  +
  O\!\left(LB\sqrt{\frac{K}{\ncal}}+
  \sqrt{\frac{\log(1/\delta)}{\ncal}}\right).
  \label{eq:parametric-bound}
\end{equation}
\end{proposition}

\paragraph{Proof of Theorem~\ref{thm:validation-selection}.}
Hoeffding's inequality and a union bound give
\(\sup_{c\in\mathcal C}|\widehat{\risk}_{\mathrm{val}}(\widehat f_c)-
\risk(\widehat f_c)|\le\epsilon\) with
\(\epsilon=\sqrt{\log(2|\mathcal C|/\delta)/(2\nval)}\). The validation argmin then
satisfies
\(\risk(\widehat f_{\widehat c})\le
\widehat{\risk}_{\mathrm{val}}(\widehat f_{\widehat c})+\epsilon
\le \widehat{\risk}_{\mathrm{val}}(\widehat f_c)+\epsilon
\le \risk(\widehat f_c)+2\epsilon\) for every \(c\in\mathcal C\).

\paragraph{Proof of Proposition~\ref{prop:table}.}
Condition on the calibration judge-output patterns, and hence on the occupied random
support \(\widehat{\mathcal Z}_K\), the counts \(N_z\), and the random size \(S_K\).
Given this conditioning, the calibration labels within each occupied cell are
independent bounded observations with mean \(g^\star(z)\), and the occupied index
set is fixed. Hoeffding's inequality and a union bound over these fixed occupied
cells give the first display with conditional probability at least \(1-\delta\).
Since the bound holds for every realized support after conditioning, the same
probability statement also holds unconditionally by integrating over the calibration
patterns. For squared loss,
\(\risk(\widetilde g)-\risk(g^\star)=
\mathbb E[(\widetilde g(Z_K)-g^\star(Z_K))^2]\) on any fixed deployed predictor. On
occupied cells,
\(|\widetilde g(z)-g^\star(z)|\le
|\bar Y_z-g^\star(z)|+\alpha/(N_z+\alpha)\), and
\((a+b)^2\le 2a^2+2b^2\). On unseen cells, both predictions lie in \([0,1]\), so the
squared contribution is at most the unseen probability \(u_K\). Integrating these
cellwise bounds with respect to the test-time probabilities \(p_z\) gives the
weighted terms in Eq.~\eqref{eq:table-integrated-bound}.

\paragraph{Proof sketch of Proposition~\ref{prop:balanced-table}.}
The lower balance condition and a Chernoff bound imply
\(N_z\ge \ncal/(2cM_K)\) for all cells with probability at least \(1-\delta\) once
\(\ncal\) is a sufficiently large multiple of \(cM_K\log(M_K/\delta)\). Substituting
this count lower bound and \(p_z\le c/M_K\) into
Eq.~\eqref{eq:table-integrated-bound} gives
Eq.~\eqref{eq:balanced-table-rate}; the same event sets the unseen-mass term to zero.
The lower bound is the standard finite-dimensional mean-estimation lower bound:
under balanced sampling, estimating \(M_K\) unrelated conditional means from
\(\ncal\) labels incurs total weighted variance of order \(M_K/\ncal\).

\paragraph{Proof sketch of Proposition~\ref{prop:stacking}.}
The empirical Rademacher complexity of bounded linear functions on \([0,1]^K\) with
\(\|w\|_2\le B\) is \(O(B\sqrt{K/\ncal})\). The contraction inequality transfers this
rate through the Lipschitz link and an \(L\)-Lipschitz loss, and the standard uniform
convergence bound gives Eq.~\eqref{eq:parametric-bound}. Ridge and logistic stacking are regularized
implementations of this low-dimensional family.

\section{Extended Empirical Results}
\label{app:extended-results}

\paragraph{Dataset construction.}
RewardBench uses the first 1000 source preference rows from
\path{allenai/reward-bench}; each chosen--rejected pair is expanded into original
and swapped orders, giving 2000 position-balanced pairwise rows. LLMBar uses all five
official subsets (Natural, Neighbor, GPTInst, GPTOut, and Manual), likewise expanded
to original and swapped orders, giving 838 rows. SummEval uses the test split of
\path{mteb/summeval}; each machine summary is labeled by the human coherence score
normalized from the original \(1\)--\(5\) rubric to \([0,1]\), giving 1600 scalar
rows. Arena100K uses \path{lmarena-ai/arena-human-preference-100k}: we keep English
non-tie rows, skip examples whose user plus two assistant conversations exceed an
estimated 5000 tokens, retain 800 source conversations, and create original and
swapped pairwise rows for 1600 total rows. For all position-balanced pairwise
datasets, grouped splits use the original source id (\(\texttt{id}\),
\(\texttt{question\_id}\), or constructed LLMBar id) so swapped copies of the same
comparison cannot cross selection, calibration, validation, and test blocks.

\paragraph{SummEval discretization.}
SummEval labels are the human coherence scores normalized from the original
\(1\)--\(5\) rubric to \([0,1]\). Judge outputs for SummEval use the direct scalar
schema: each judge is prompted for a score in \(\{1,2,3,4,5\}\), and any parsed
numeric response is rounded and clipped to that range before entering the joint
pattern table. Thus table sparsity on SummEval is driven by products of five-level
judge bins rather than pairwise verdict alphabets. Table~\ref{tab:table-sensitivity}
checks that the largest-budget joint-table conclusion is stable across smoothing and
fallback choices; sensitivity to alternative rubric binnings remains a natural
extension.

\paragraph{Judge configuration and deterministic outputs.}
The fixed judge pool uses Qwen2.5-7B-Instruct \citep{qwen2024qwen25},
Llama-3.1-8B-Instruct \citep{grattafiori2024llama3}, Mistral-7B-Instruct-v0.3
\citep{jiang2023mistral}, Prometheus-7B-v2.0 \citep{kim2024prometheus2},
Gemma-3-12B-IT \citep{gemmateam2025gemma3}, Selene-1-Mini-Llama-3.1-8B
\citep{alexandru2025selene}, and DeepSeek V4 Flash \citep{deepseekai2026deepseekv4}.
All seven judge outputs use a direct schema: pairwise tasks request a JSON verdict in
\(\{A,B,\mathrm{tie}\}\), and SummEval requests a JSON score in \(\{1,\ldots,5\}\). Local vLLM and
transformers runs use deterministic decoding with temperature \(0\) or
\(\texttt{do\_sample=False}\); the DeepSeek API anchor uses temperature \(0\), JSON
response formatting, and non-thinking mode. We treat each judge as a deterministic
annotator for this paper. Repeated stochastic samples per judge would create a
different candidate family and are left to jury/repeated-sample methods; the
sensitivity checks below instead vary pool membership and downstream calibration
choices around this fixed measurement protocol.

\paragraph{Prompt and parsing details.}
For generic pairwise judges, the system message is
``You are a strict preference judge. Return only valid JSON,'' and the user prompt
asks which response is better for the original user prompt, preferring helpful,
correct, harmless, and instruction-following answers, then requires exactly
\(\{\texttt{``verdict''}:\texttt{``A''}\mid\texttt{``B''}\mid\texttt{``tie''}\}\).
For SummEval-style scalar judges, the system message is
``You are a strict summary-evaluation judge. Return only valid JSON,'' and the user
prompt asks for an integer \(1\)--\(5\) score on the requested dimension, with
1 meaning very poor and 5 excellent. Prometheus uses its recommended feedback-style
prompt with a final \texttt{[RESULT]} marker, but the parsed output is collapsed to
the same verdict or score alphabet. Local runs use \(\texttt{max\_tokens}=80\) for
generic prompts and 128--512 for Prometheus-style feedback prompts; DeepSeek uses
\(\texttt{max\_tokens}=80\), JSON response formatting, and up to three retries under
the same schema. Parsed scalar scores are rounded and
clipped to \(\{1,\ldots,5\}\); invalid categorical verdicts fall back to tie, while
unparseable outputs are recorded as \(\texttt{parse\_error}\).
When these categorical pairwise outputs are used by scalar aggregators, \(A/a\) is
mapped to 1, \(B/b\) to 0, and both \(\texttt{tie}\) and
\(\texttt{parse\_error}\) to \(1/2\); the joint-table family instead treats each
recorded output category as a distinct cell value.
For a clean prompt-template perturbation, Table~\ref{tab:prompt-perturbation}
uses the four general local pairwise judges on LLMBar. DeepSeek is API-only and
therefore belongs to the pool-perturbation check, Prometheus is evaluated in its
feedback-style format rather than forced into a letter-only template, and SummEval
uses a scalar score schema rather than pairwise \(\texttt{A}/\texttt{B}/\texttt{tie}\)
verdicts.

\begin{table}[t]
\centering
\small
\begin{tabular}{p{0.18\linewidth}p{0.36\linewidth}p{0.36\linewidth}}
\toprule
Block & Used for & Not used for \\
\midrule
Selection block & Construct judge paths and path-specific oracle diagnostics & Fitting calibration maps or reporting final risk \\
Calibration block & Fit table, stacking, one-coin, and calibration maps & Choosing the final candidate by test performance \\
Validation block & Select \(K\), path, and/or aggregator family within a split & Estimating final held-out performance \\
Test block & Report final risk and diagnostics after selection is fixed & Choosing budget, \(K\), path, aggregator, smoothing, or fallback \\
\bottomrule
\end{tabular}
\caption{Protocol guardrails. All deployed choices are made before the test block is evaluated. Tables that compare budgets or families report either fixed largest-budget results or validation-selected choices, avoiding test-informed model selection.}
\label{tab:protocol-guardrails}
\end{table}

\begin{table}[t]
\centering
\small
\begin{tabular}{lccccc}
\toprule
Dataset & $\ncal$ & six-pool $K/m$ & six-pool MSE & seven-pool $K/m$ & seven-pool MSE \\
\midrule
RewardBench & 800 & 3.40/6 & 0.026 & 3.37/7 & 0.024 \\
LLMBar & 300 & 4.17/6 & 0.226 & 4.47/7 & 0.204 \\
SummEval & 800 & 1.77/6 & 0.045 & 2.43/7 & 0.044 \\
Arena100K & 400 & 2.80/6 & 0.242 & 2.50/7 & 0.231 \\
\bottomrule
\end{tabular}
\caption{Judge-pool perturbation check at the largest calibrator-fitting budget on a fixed information-first path. The six-judge pool contains Qwen, Llama, Mistral, Prometheus, Gemma, and Selene; the seven-judge pool adds DeepSeek V4 Flash. Values are split means. Adding a strong API judge changes risk levels, especially on LLMBar and Arena100K, but selected prefixes remain below the full pool in all four datasets.}
\label{tab:pool-artifact-guardrail}
\end{table}

\begin{table}[t]
\centering
\scriptsize
\setlength{\tabcolsep}{3pt}
\begin{tabular}{lccccccc}
\toprule
$\ncal$ & JSON $K/4$ & JSON & $u$ & Letter $K/4$ & Letter & $u$ & $\Delta$ \\
\midrule
20 & 1.93/4 & 0.250 & 0.056 & 2.13/4 & 0.256 & 0.150 & +0.006 \\
50 & 2.20/4 & 0.244 & 0.033 & 1.60/4 & 0.248 & 0.020 & +0.004 \\
100 & 2.60/4 & 0.235 & 0.027 & 2.03/4 & 0.243 & 0.020 & +0.008 \\
200 & 3.13/4 & 0.231 & 0.020 & 2.03/4 & 0.234 & 0.009 & +0.003 \\
300 & 3.13/4 & 0.225 & 0.017 & 2.60/4 & 0.229 & 0.010 & +0.004 \\
\bottomrule
\end{tabular}
\caption{Prompt-perturbation check on LLMBar with four general local judges (Qwen, Llama, Mistral, and Selene). JSON is the main direct schema; Letter asks only for \texttt{A}, \texttt{B}, or \texttt{tie}. The prompt change alters 32.9\% of individual judge verdicts on average, with parse-error rates 0.03\%/ 0.06\% for JSON/Letter; selected prefixes remain below the full pool under both prompt schemes.}
\label{tab:prompt-perturbation}
\end{table}

\begin{table}[t]
\centering
\small
\begin{tabular}{lccccc}
\toprule
Dataset & \(n_V\) & table menu & table scale & scalar menu & scalar scale \\
\midrule
RewardBench & 300 & \(3\times 7=21\) & 0.101 & \(6\times 7=42\) & 0.112 \\
LLMBar & 100 & \(3\times 7=21\) & 0.174 & \(6\times 7=42\) & 0.193 \\
SummEval & 192 & \(3\times 7=21\) & 0.126 & \(6\times 7=42\) & 0.139 \\
Arena100K & 200 & \(3\times 7=21\) & 0.123 & \(6\times 7=42\) & 0.137 \\
\bottomrule
\end{tabular}
\caption{Candidate-menu sizes and validation-burden scales in the main real-data comparisons. The joint-table menu validates over three path rules and seven prefix sizes. The scalar menu validates over six scalar/reliability/stacking families and seven prefix sizes. The scale columns report \(\sqrt{\log|\mathcal C|/n_V}\), the menu-dependent part of the finite-validation term. The fixed-\(\nhuman\) allocation diagnostic uses a compact one-path menu of \(7\times(6+1)=49\) scalar-or-table candidates.}
\label{tab:candidate-menu-validation}
\end{table}

\begin{table}[t]
\centering
\small
\begin{tabular}{p{0.28\linewidth}p{0.33\linewidth}p{0.29\linewidth}}
\toprule
Theory component & Empirical evidence & Interpretation \\
\midrule
Validation selector over \((K,\aggfam)\) & Tables~\ref{tab:baseline-envelope}--\ref{tab:regime-comparison} & Selection is evaluated with fixed-budget validation and held-out test evaluation. \\
Joint-table support term \(H_K/\ncal\) and unseen mass & Figures~\ref{fig:risk-k}, \ref{fig:complexity-risk}, \ref{fig:unseen-risk}; Table~\ref{tab:controlled-complexity} & Prefix figures visualize finite-sample behavior inside the joint-table family. \\
Operational table-pressure instantiation & Table~\ref{tab:exact-cell-pressure} & Calibration cell counts and held-out pattern frequencies instantiate the plug-in cell-pressure term \(\sum_z\widehat p_z/N_z+\widehat u\), with \(H_K/n_M+\widehat u\) as a compact proxy. \\
Low-dimensional/reliability aggregators plus approximation bias & Tables~\ref{tab:aggregation-baselines}--\ref{tab:metric-robustness}; Table~\ref{tab:regime-comparison} & Scalar aggregation and one-coin reliability models often win on real benchmarks because they avoid the joint pattern table, and this is checked across MSE, NLL, accuracy, and correlation metrics. \\
Theory proxies for family choice & Table~\ref{tab:theory-proxy-winner} & \(H_K/n_M\), unseen rate, and pairwise gain provide exploratory regime signals for the validation selector. \\
Smoothing and fallback terms in the table bound & Table~\ref{tab:table-sensitivity} & Joint-table conclusions are checked against alternative smoothing constants and lower-prefix fallback. \\
Approximation--estimation regime distinction & Table~\ref{tab:controlled-aggregator-regimes} & Additive regimes favor stacking; interaction regimes can favor the joint table when labels are sufficient. \\
\bottomrule
\end{tabular}
\caption{How the \((K,\aggfam)\) theory is supported empirically. Prefix diagnostics describe the joint-table family; aggregator-level evidence comes from scalar baselines, regime comparisons, and controlled additive/interaction experiments.}
\label{tab:theory-evidence-map}
\end{table}

\begin{table}[t]
\centering
\scriptsize
\begin{tabular}{@{}p{0.21\linewidth}p{0.35\linewidth}p{0.29\linewidth}@{}}
\toprule
Interpretation question & Evidence check & Takeaway \\
\midrule
What does validation select? &
Theorem~\ref{thm:validation-selection}; Tables~\ref{tab:theory-evidence-map} and \ref{tab:k-family-ablation} &
Validation is the selection primitive; the contribution is the finite-calibration candidate space, diagnostics, and regime evidence. \\
Does \(K\)-selection matter beyond family choice? &
Table~\ref{tab:k-family-ablation}; Figure~\ref{fig:marginal} &
Family choice is the larger real-data effect; prefix selection mainly protects joint tables from all-seven support pressure. \\
Are conclusions MSE-specific? &
Tables~\ref{tab:metric-robustness} and \ref{tab:nll-selection-ablation} &
The scalar/reliability pattern also appears under non-MSE metrics, with MSE as the theory-aligned objective. \\
Are there middle families between scalars and full tables? &
Tables~\ref{tab:backoff-split-sensitivity} and \ref{tab:residual-shrinkage-hybrid} &
Backoff and residual-shrinkage variants are diagnostic middle points with limited gains over the strongest scalar selector. \\
Is the path heuristic essential? &
Table~\ref{tab:cfps-path}; Figure~\ref{fig:complexity-risk-path} &
Path choice affects some cells, while the prefix/family decision remains the main finite-calibration object. \\
How should table diagnostics be read? &
Table~\ref{tab:exact-cell-pressure}; Table~\ref{tab:theory-proxy-winner}; Figures~\ref{fig:complexity-risk}--\ref{fig:unseen-risk} &
\(H_K/n_M\) is an exploratory pressure proxy; exact cell pressure and unseen mass provide the closer audit quantities. \\
Does SummEval binning drive the result? &
Appendix ``SummEval discretization''; Table~\ref{tab:table-sensitivity} &
The fixed five-bin schema is documented, smoothing/fallback sensitivity is checked, and alternative binnings remain future work. \\
How far does the fixed judge pool generalize? &
Appendix ``Judge configuration and deterministic outputs''; Limitations &
The study covers a fixed deployed pool; larger heterogeneous and stochastic pools are future work. \\
\bottomrule
\end{tabular}
\caption{Navigation map from interpretation questions to the corresponding evidence checks in the draft. Several checks are regime diagnostics that connect the theory terms to observed finite-label behavior.}
\label{tab:interpretation-checks}
\end{table}

\begin{table}[!htbp]
\centering
\small
\begin{tabular}{lcccc}
\toprule
Method & RewardBench & LLMBar & SummEval & Arena100K \\
\midrule
Best single judge (val.) & 0.032 $\pm$ 0.001 & 0.219 $\pm$ 0.004 & 0.045 $\pm$ 0.001 & 0.234 $\pm$ 0.002 \\
All seven judges & 0.028 $\pm$ 0.001 & 0.211 $\pm$ 0.002 & 0.059 $\pm$ 0.001 & 0.245 $\pm$ 0.002 \\
Information-first + val. $K$ & 0.024 $\pm$ 0.001 & 0.204 $\pm$ 0.002 & 0.044 $\pm$ 0.001 & 0.231 $\pm$ 0.002 \\
Complexity-penalized + val. $K$ & 0.025 $\pm$ 0.001 & 0.204 $\pm$ 0.002 & 0.046 $\pm$ 0.001 & 0.232 $\pm$ 0.002 \\
Random path + val. $K$ & 0.027 $\pm$ 0.001 & 0.207 $\pm$ 0.003 & 0.048 $\pm$ 0.001 & 0.240 $\pm$ 0.002 \\
Path-family envelope & 0.025 $\pm$ 0.001 & 0.203 $\pm$ 0.002 & 0.045 $\pm$ 0.001 & 0.233 $\pm$ 0.002 \\
Test oracle envelope & 0.022 $\pm$ 0.001 & 0.194 $\pm$ 0.002 & 0.043 $\pm$ 0.001 & 0.227 $\pm$ 0.002 \\
\bottomrule
\end{tabular}
\caption{Baseline and envelope test risk at each dataset's largest available calibrator-fitting budget. Values are mean $\pm$ standard error over splits. The path-family envelope selects the path rule and prefix size by validation risk within each split and fixed budget; the test oracle envelope is a non-deployable diagnostic upper bound. No calibrator-fitting budget is selected using test risk.}
\label{tab:baseline-envelope}
\end{table}

\begin{table}[t]
\centering
\small
\begin{tabular}{lccc}
\toprule
Dataset & reporting budget $n_M$ & envelope risk & envelope mean $K$ \\
\midrule
RewardBench & 800 & 0.025 $\pm$ 0.001 & 3.63 \\
LLMBar & 300 & 0.203 $\pm$ 0.002 & 4.13 \\
SummEval & 800 & 0.045 $\pm$ 0.001 & 2.47 \\
Arena100K & 400 & 0.233 $\pm$ 0.002 & 2.73 \\
\bottomrule
\end{tabular}
\caption{Fixed reporting budget and selected complexity for the validation-selected path-family envelope used in Table~\ref{tab:baseline-envelope}. The reporting budget is the largest available calibrator-fitting budget for each dataset.}
\label{tab:envelope-budget-k}
\end{table}

\begin{table}[t]
\centering
\small
\begin{tabular}{lcccc}
\toprule
Method & RewardBench & LLMBar & SummEval & Arena100K \\
\midrule
All-judge mean + isotonic & 0.022 $\pm$ 0.001 & 0.245 $\pm$ 0.002 & 0.047 $\pm$ 0.001 & 0.231 $\pm$ 0.001 \\
One-coin reliability + isotonic & 0.021 $\pm$ 0.001 & 0.195 $\pm$ 0.003 & 0.050 $\pm$ 0.001 & 0.229 $\pm$ 0.002 \\
Ridge stacking + isotonic & 0.023 $\pm$ 0.001 & 0.184 $\pm$ 0.002 & 0.043 $\pm$ 0.001 & 0.231 $\pm$ 0.002 \\
Logistic stacking & 0.021 $\pm$ 0.001 & 0.181 $\pm$ 0.002 & 0.059 $\pm$ 0.001 & 0.228 $\pm$ 0.002 \\
Ridge pairwise stacking + isotonic & 0.025 $\pm$ 0.001 & 0.192 $\pm$ 0.002 & 0.044 $\pm$ 0.001 & 0.237 $\pm$ 0.002 \\
Logistic pairwise stacking & 0.021 $\pm$ 0.001 & 0.183 $\pm$ 0.002 & 0.060 $\pm$ 0.001 & 0.229 $\pm$ 0.002 \\
\bottomrule
\end{tabular}
\caption{Scalar aggregation baselines at each dataset's largest calibrator-fitting budget. All methods use all seven judge outputs and avoid the full joint pattern table. Pairwise verdicts are encoded as \(A/a=1\), \(B/b=0\), and tie or parse-error outcomes as \(1/2\) before fitting scalar models. The reported metric is mean squared calibration error, matching Table~\ref{tab:baseline-envelope}.}
\label{tab:aggregation-baselines}
\end{table}

\begin{table}[t]
\centering
\small
\begin{tabular}{lcccc}
\toprule
Dataset & $n_M$ & selected aggregation family & selected $K$ & test MSE \\
\midrule
RewardBench & 800 & Logistic pairwise (0.23) & 6.17 & 0.023 $\pm$ 0.001 \\
LLMBar & 300 & Logistic (0.43) & 7.00 & 0.183 $\pm$ 0.002 \\
SummEval & 800 & Ridge + isotonic (0.63) & 6.70 & 0.044 $\pm$ 0.001 \\
Arena100K & 400 & One-coin reliability + isotonic (0.43) & 6.53 & 0.232 $\pm$ 0.002 \\
\bottomrule
\end{tabular}
\caption{Validation-selected scalar aggregation at each dataset's largest calibrator-fitting budget. The selector chooses both aggregation family and prefix size using validation risk within each split; the parenthesized value is the split share of the most frequently selected family.}
\label{tab:aggregation-selected}
\end{table}

\begin{table}[t]
\centering
\scriptsize
\resizebox{\linewidth}{!}{%
\begin{tabular}{lccccccc}
\toprule
Dataset & \(n_M\) & table \(K{=}7\) & table sel. \(K\) & scalar \(K{=}7\) & scalar sel. \(K\) & full \(K{=}7\) & full sel. \(K\) \\
\midrule
RewardBench & 800 & 0.0277 & 0.0251 & 0.0218 & 0.0225 & 0.0218 & 0.0234 \\
LLMBar & 300 & 0.2110 & 0.2031 & 0.1857 & 0.1857 & 0.1863 & 0.1886 \\
SummEval & 800 & 0.0588 & 0.0454 & 0.0433 & 0.0435 & 0.0433 & 0.0436 \\
Arena100K & 400 & 0.2448 & 0.2331 & 0.2316 & 0.2324 & 0.2318 & 0.2318 \\
\bottomrule
\end{tabular}%
}
\caption{\(K\)-versus-family ablation at each dataset's largest calibrator-fitting budget.
All entries are split-wise validation selectors over existing candidates, reported as
mean test MSE. ``table \(K{=}7\)'' fixes the all-seven joint table; ``table sel.
\(K\)'' selects the table prefix by validation risk; ``scalar \(K{=}7\)'' selects
the scalar family at the all-seven prefix; ``scalar sel. \(K\)'' selects both scalar
family and prefix; ``full'' pools joint-table and scalar candidates. The main real-data
gain comes from choosing an estimable aggregation family rather than from scalar-family
prefix selection alone, while prefix selection protects the joint table from all-seven
support pressure.}
\label{tab:k-family-ablation}
\end{table}

\begin{table}[t]
\centering
\scriptsize
\resizebox{\linewidth}{!}{%
\begin{tabular}{lcccccc}
\toprule
Dataset & $n_M$ & selected $K$ mean $\pm$ sd & modal $K$ share & test MSE & regret vs. $K\le 3$ & oracle gap \\
\midrule
RewardBench & 50 & 2.57 $\pm$ 0.68 & 2 (0.53) & 0.026 $\pm$ 0.001 & +0.000 $\pm$ 0.000 & 0.002 $\pm$ 0.001 \\
 & 100 & 2.83 $\pm$ 1.02 & 3 (0.37) & 0.025 $\pm$ 0.001 & +0.000 $\pm$ 0.000 & 0.002 $\pm$ 0.000 \\
 & 200 & 2.87 $\pm$ 0.90 & 3 (0.40) & 0.025 $\pm$ 0.001 & +0.000 $\pm$ 0.000 & 0.002 $\pm$ 0.000 \\
 & 400 & 3.20 $\pm$ 1.30 & 4 (0.30) & 0.025 $\pm$ 0.001 & -0.000 $\pm$ 0.001 & 0.004 $\pm$ 0.001 \\
 & 800 & 3.63 $\pm$ 1.25 & 3 (0.37) & 0.025 $\pm$ 0.001 & -0.001 $\pm$ 0.001 & 0.003 $\pm$ 0.001 \\
\addlinespace
LLMBar & 20 & 2.37 $\pm$ 1.47 & 1 (0.40) & 0.226 $\pm$ 0.003 & -0.001 $\pm$ 0.001 & 0.007 $\pm$ 0.002 \\
 & 50 & 3.03 $\pm$ 1.40 & 2 (0.27) & 0.222 $\pm$ 0.003 & -0.001 $\pm$ 0.002 & 0.014 $\pm$ 0.003 \\
 & 100 & 3.30 $\pm$ 1.26 & 4 (0.30) & 0.214 $\pm$ 0.003 & +0.000 $\pm$ 0.002 & 0.011 $\pm$ 0.002 \\
 & 200 & 4.00 $\pm$ 1.68 & 3 (0.23) & 0.214 $\pm$ 0.004 & +0.006 $\pm$ 0.003 & 0.016 $\pm$ 0.003 \\
 & 300 & 4.13 $\pm$ 1.33 & 3 (0.27) & 0.203 $\pm$ 0.002 & -0.003 $\pm$ 0.002 & 0.009 $\pm$ 0.001 \\
\addlinespace
SummEval & 32 & 1.47 $\pm$ 0.68 & 1 (0.63) & 0.056 $\pm$ 0.001 & +0.000 $\pm$ 0.000 & 0.002 $\pm$ 0.000 \\
 & 64 & 1.80 $\pm$ 0.85 & 1 (0.47) & 0.053 $\pm$ 0.001 & +0.000 $\pm$ 0.000 & 0.002 $\pm$ 0.001 \\
 & 128 & 2.10 $\pm$ 1.03 & 2 (0.43) & 0.050 $\pm$ 0.001 & +0.001 $\pm$ 0.001 & 0.003 $\pm$ 0.001 \\
 & 256 & 1.87 $\pm$ 0.68 & 2 (0.53) & 0.046 $\pm$ 0.001 & +0.000 $\pm$ 0.000 & 0.002 $\pm$ 0.001 \\
 & 512 & 2.33 $\pm$ 0.96 & 2 (0.57) & 0.045 $\pm$ 0.001 & +0.000 $\pm$ 0.000 & 0.002 $\pm$ 0.000 \\
 & 800 & 2.47 $\pm$ 1.11 & 2 (0.37) & 0.045 $\pm$ 0.001 & +0.001 $\pm$ 0.001 & 0.003 $\pm$ 0.001 \\
\addlinespace
Arena100K & 50 & 2.43 $\pm$ 1.57 & 1 (0.33) & 0.239 $\pm$ 0.002 & +0.002 $\pm$ 0.001 & 0.005 $\pm$ 0.001 \\
 & 100 & 2.30 $\pm$ 1.56 & 1 (0.40) & 0.236 $\pm$ 0.003 & +0.002 $\pm$ 0.001 & 0.006 $\pm$ 0.001 \\
 & 200 & 2.33 $\pm$ 1.32 & 2 (0.33) & 0.234 $\pm$ 0.002 & +0.002 $\pm$ 0.001 & 0.006 $\pm$ 0.001 \\
 & 400 & 2.73 $\pm$ 1.31 & 3 (0.27) & 0.233 $\pm$ 0.002 & +0.000 $\pm$ 0.001 & 0.006 $\pm$ 0.002 \\
\bottomrule
\end{tabular}%
}
\caption{Selected-prefix instability and regret for the joint-table selector across calibrator-fitting budgets. Selection is split-wise validation over the deployable non-oracle table candidates. The regret column compares the selected table to a deployable short-prefix rule that uses the same validation criterion but restricts to $K\le 3$; positive values mean the unrestricted selected-$K$ table is worse. The oracle gap is the non-deployable difference from the best table candidate on the held-out test split.}
\label{tab:selected-k-instability}
\end{table}

\begin{table}[t]
\centering
\scriptsize
\begin{tabular}{lrrrrr}
\toprule
Dataset & $n_M$ & table & scalar & $\Delta$ & paired 95\% CI \\
\midrule
RewardBench & 50 & 0.0252 & 0.0219 & 0.0033 & [0.0013,0.0053] \\
RewardBench & 100 & 0.0249 & 0.0225 & 0.0025 & [0.0012,0.0038] \\
RewardBench & 200 & 0.0251 & 0.0220 & 0.0032 & [0.0017,0.0046] \\
RewardBench & 400 & 0.0243 & 0.0217 & 0.0026 & [0.0015,0.0037] \\
RewardBench & 800 & 0.0240 & 0.0227 & 0.0014 & [-0.0001,0.0028] \\
LLMBar & 20 & 0.2254 & 0.2212 & 0.0042 & [-0.0036,0.0120] \\
LLMBar & 50 & 0.2213 & 0.2111 & 0.0102 & [0.0033,0.0171] \\
LLMBar & 100 & 0.2147 & 0.1972 & 0.0176 & [0.0126,0.0226] \\
LLMBar & 200 & 0.2104 & 0.1914 & 0.0189 & [0.0123,0.0256] \\
LLMBar & 300 & 0.2039 & 0.1826 & 0.0214 & [0.0186,0.0241] \\
SummEval & 32 & 0.0569 & 0.0511 & 0.0058 & [0.0032,0.0084] \\
SummEval & 64 & 0.0528 & 0.0501 & 0.0026 & [0.0004,0.0048] \\
SummEval & 128 & 0.0485 & 0.0474 & 0.0011 & [-0.0006,0.0028] \\
SummEval & 256 & 0.0451 & 0.0460 & -0.0009 & [-0.0022,0.0005] \\
SummEval & 512 & 0.0448 & 0.0436 & 0.0011 & [0.0002,0.0021] \\
SummEval & 800 & 0.0444 & 0.0435 & 0.0009 & [-0.0001,0.0019] \\
Arena100K & 50 & 0.2383 & 0.2372 & 0.0011 & [-0.0033,0.0056] \\
Arena100K & 100 & 0.2326 & 0.2349 & -0.0023 & [-0.0056,0.0010] \\
Arena100K & 200 & 0.2328 & 0.2331 & -0.0003 & [-0.0035,0.0029] \\
Arena100K & 400 & 0.2311 & 0.2321 & -0.0010 & [-0.0045,0.0026] \\
\bottomrule
\end{tabular}
\caption{Cross-budget regime comparison. The table column uses the lowest-risk validation-selected path rule within the joint-table family; the scalar column uses validation to select scalar aggregation family and prefix size at the same budget. Here $\Delta$ is table risk minus scalar risk, so positive values favor scalar aggregation. Paired intervals are normal 95\% intervals over the 30 split-wise differences for the same fixed table rule and scalar selector.}
\label{tab:regime-comparison}
\end{table}

\begin{table}[t]
\centering
\scriptsize
\resizebox{\linewidth}{!}{%
\begin{tabular}{lrrrrrr}
\toprule
Dataset & $n_M$ & scalar oracle & table oracle & scalar finite & table finite & finite/oracle reading \\
\midrule
Arena100K & 50 & 0.232 $\pm$ 0.002 & 0.234 $\pm$ 0.002 & 0.237 $\pm$ 0.002 & 0.239 $\pm$ 0.002 & no table oracle gain \\
Arena100K & 100 & 0.229 $\pm$ 0.002 & 0.229 $\pm$ 0.002 & 0.235 $\pm$ 0.002 & 0.236 $\pm$ 0.003 & no table oracle gain \\
Arena100K & 200 & 0.228 $\pm$ 0.002 & 0.228 $\pm$ 0.002 & 0.233 $\pm$ 0.002 & 0.234 $\pm$ 0.002 & no table oracle gain \\
Arena100K & 400 & 0.226 $\pm$ 0.001 & 0.227 $\pm$ 0.002 & 0.232 $\pm$ 0.002 & 0.233 $\pm$ 0.002 & no table oracle gain \\
LLMBar & 20 & 0.215 $\pm$ 0.003 & 0.219 $\pm$ 0.003 & 0.221 $\pm$ 0.004 & 0.226 $\pm$ 0.003 & no table oracle gain \\
LLMBar & 50 & 0.200 $\pm$ 0.004 & 0.208 $\pm$ 0.003 & 0.211 $\pm$ 0.004 & 0.222 $\pm$ 0.003 & no table oracle gain \\
LLMBar & 100 & 0.193 $\pm$ 0.003 & 0.203 $\pm$ 0.002 & 0.197 $\pm$ 0.003 & 0.214 $\pm$ 0.003 & no table oracle gain \\
LLMBar & 200 & 0.181 $\pm$ 0.002 & 0.197 $\pm$ 0.003 & 0.191 $\pm$ 0.004 & 0.214 $\pm$ 0.004 & no table oracle gain \\
LLMBar & 300 & 0.179 $\pm$ 0.002 & 0.194 $\pm$ 0.002 & 0.183 $\pm$ 0.002 & 0.203 $\pm$ 0.002 & no table oracle gain \\
RewardBench & 50 & 0.019 $\pm$ 0.001 & 0.024 $\pm$ 0.001 & 0.022 $\pm$ 0.001 & 0.026 $\pm$ 0.001 & no table oracle gain \\
RewardBench & 100 & 0.020 $\pm$ 0.001 & 0.023 $\pm$ 0.001 & 0.022 $\pm$ 0.001 & 0.025 $\pm$ 0.001 & no table oracle gain \\
RewardBench & 200 & 0.020 $\pm$ 0.001 & 0.023 $\pm$ 0.001 & 0.022 $\pm$ 0.001 & 0.025 $\pm$ 0.001 & no table oracle gain \\
RewardBench & 400 & 0.019 $\pm$ 0.001 & 0.021 $\pm$ 0.001 & 0.022 $\pm$ 0.001 & 0.025 $\pm$ 0.001 & no table oracle gain \\
RewardBench & 800 & 0.020 $\pm$ 0.001 & 0.022 $\pm$ 0.001 & 0.023 $\pm$ 0.001 & 0.025 $\pm$ 0.001 & no table oracle gain \\
SummEval & 32 & 0.050 $\pm$ 0.001 & 0.054 $\pm$ 0.001 & 0.051 $\pm$ 0.001 & 0.056 $\pm$ 0.001 & no table oracle gain \\
SummEval & 64 & 0.048 $\pm$ 0.001 & 0.051 $\pm$ 0.001 & 0.050 $\pm$ 0.001 & 0.053 $\pm$ 0.001 & no table oracle gain \\
SummEval & 128 & 0.046 $\pm$ 0.000 & 0.047 $\pm$ 0.001 & 0.047 $\pm$ 0.001 & 0.050 $\pm$ 0.001 & no table oracle gain \\
SummEval & 256 & 0.043 $\pm$ 0.001 & 0.044 $\pm$ 0.001 & 0.046 $\pm$ 0.001 & 0.046 $\pm$ 0.001 & no table oracle gain \\
SummEval & 512 & 0.043 $\pm$ 0.001 & 0.043 $\pm$ 0.001 & 0.044 $\pm$ 0.001 & 0.045 $\pm$ 0.001 & no table oracle gain \\
SummEval & 800 & 0.043 $\pm$ 0.001 & 0.043 $\pm$ 0.001 & 0.044 $\pm$ 0.001 & 0.045 $\pm$ 0.001 & no table oracle gain \\
\bottomrule
\end{tabular}%
}
\caption{Oracle interaction-gap audit for all real-data dataset--budget cells. The oracle columns select the lowest held-out test risk within the scalar or joint-table candidate family and are non-deployable diagnostics; finite columns select by validation risk. A cell marked no table oracle gain means the joint table does not materially beat scalar even with test-oracle selection; finite estimation means the joint-table oracle is better but validation-selected finite tables lose under sparse support or selection noise.}
\label{tab:oracle-interaction-gap-full}
\end{table}

\begin{table}[t]
\centering
\scriptsize
\resizebox{\linewidth}{!}{%
\begin{tabular}{lccccc}
\toprule
Dataset & $n_M$ & best table-like & scalar & table-like $-$ scalar & selected table-like family \\
\midrule
RewardBench & 50 & 0.026 $\pm$ 0.001 & 0.022 $\pm$ 0.001 & +0.004 [+0.002,+0.006] & joint table (0.83) \\
 & 100 & 0.024 $\pm$ 0.001 & 0.022 $\pm$ 0.001 & +0.002 [+0.000,+0.004] & joint table (0.57) \\
 & 200 & 0.025 $\pm$ 0.001 & 0.022 $\pm$ 0.001 & +0.003 [+0.002,+0.004] & joint table (0.53) \\
 & 400 & 0.024 $\pm$ 0.001 & 0.022 $\pm$ 0.001 & +0.003 [+0.001,+0.004] & joint table (0.53) \\
 & 800 & 0.023 $\pm$ 0.001 & 0.023 $\pm$ 0.001 & -0.000 [-0.001,+0.001] & residual (0.73) \\
\addlinespace
LLMBar & 20 & 0.222 $\pm$ 0.003 & 0.221 $\pm$ 0.004 & +0.001 [-0.006,+0.008] & hier. backoff (0.70) \\
 & 50 & 0.219 $\pm$ 0.003 & 0.211 $\pm$ 0.004 & +0.008 [+0.002,+0.014] & hier. backoff (0.60) \\
 & 100 & 0.214 $\pm$ 0.003 & 0.197 $\pm$ 0.003 & +0.017 [+0.012,+0.022] & joint table (0.60) \\
 & 200 & 0.210 $\pm$ 0.004 & 0.191 $\pm$ 0.004 & +0.019 [+0.010,+0.027] & joint table (0.53) \\
 & 300 & 0.186 $\pm$ 0.002 & 0.183 $\pm$ 0.002 & +0.003 [+0.000,+0.006] & residual (0.90) \\
\addlinespace
SummEval & 32 & 0.054 $\pm$ 0.001 & 0.051 $\pm$ 0.001 & +0.003 [+0.001,+0.005] & hier. backoff (0.87) \\
 & 64 & 0.053 $\pm$ 0.001 & 0.050 $\pm$ 0.001 & +0.002 [+0.000,+0.005] & hier. backoff (0.90) \\
 & 128 & 0.049 $\pm$ 0.001 & 0.047 $\pm$ 0.001 & +0.002 [+0.000,+0.004] & hier. backoff (0.67) \\
 & 256 & 0.045 $\pm$ 0.001 & 0.046 $\pm$ 0.001 & -0.001 [-0.002,+0.001] & hier. backoff (0.73) \\
 & 512 & 0.044 $\pm$ 0.001 & 0.044 $\pm$ 0.001 & +0.001 [-0.000,+0.002] & hier. backoff (0.63) \\
 & 800 & 0.044 $\pm$ 0.001 & 0.044 $\pm$ 0.001 & +0.000 [-0.000,+0.001] & residual (0.53) \\
\addlinespace
Arena100K & 50 & 0.235 $\pm$ 0.002 & 0.237 $\pm$ 0.002 & -0.002 [-0.006,+0.001] & hier. backoff (0.63) \\
 & 100 & 0.232 $\pm$ 0.002 & 0.235 $\pm$ 0.002 & -0.002 [-0.006,+0.001] & hier. backoff (0.80) \\
 & 200 & 0.232 $\pm$ 0.002 & 0.233 $\pm$ 0.002 & -0.001 [-0.004,+0.003] & hier. backoff (0.77) \\
 & 400 & 0.232 $\pm$ 0.002 & 0.232 $\pm$ 0.002 & -0.000 [-0.003,+0.002] & residual (0.40) \\
\bottomrule
\end{tabular}%
}
\caption{Cross-budget comparison against a validation-selected table-like menu. For each split and budget, the table-like side validates over the unrestricted joint table and hierarchical backoff table; at the largest budgets it also includes the count-shrunk residual table from Table~\ref{tab:residual-shrinkage-hybrid}. It is then compared on the same held-out split to the scalar-family selector. Positive gaps favor scalar aggregation. This diagnostic places unrestricted and shrunk table-like estimators on the same validation-selection scale as the scalar-family selector.}
\label{tab:table-like-family-comparison}
\end{table}

\begin{table}[t]
\centering
\small
\begin{tabular}{lccccc}
\toprule
Dataset & MSE & NLL & Accuracy & Pearson & Spearman \\
\midrule
RewardBench & 0.023 $\pm$ 0.001 & 0.101 $\pm$ 0.005 & 0.971 $\pm$ 0.001 & 0.954 $\pm$ 0.002 & 0.914 $\pm$ 0.003 \\
LLMBar & 0.183 $\pm$ 0.002 & 0.557 $\pm$ 0.010 & 0.719 $\pm$ 0.006 & 0.522 $\pm$ 0.008 & 0.522 $\pm$ 0.008 \\
SummEval & 0.044 $\pm$ 0.001 & 0.621 $\pm$ 0.002 & 0.763 $\pm$ 0.004 & 0.601 $\pm$ 0.007 & 0.592 $\pm$ 0.007 \\
Arena100K & 0.232 $\pm$ 0.002 & 0.665 $\pm$ 0.006 & 0.626 $\pm$ 0.006 & 0.273 $\pm$ 0.012 & 0.273 $\pm$ 0.013 \\
\bottomrule
\end{tabular}
\caption{Metric robustness for validation-selected scalar aggregation at the largest calibrator-fitting budget. MSE is the main paper metric; NLL, accuracy, Pearson correlation, and Spearman correlation show the same scalar-aggregation regime across complementary evaluation criteria.}
\label{tab:metric-robustness}
\end{table}

\begin{table}[!htbp]
\centering
\small
\begin{tabular}{lcccc}
\toprule
Dataset & joint table & scalar selector & DS/BT-style jury & selected $K$ \\
\midrule
Arena100K & 0.233 & 0.232 & 0.235 $\pm$ 0.002 & 6.07 \\
LLMBar & 0.203 & 0.183 & 0.238 $\pm$ 0.002 & 2.33 \\
RewardBench & 0.025 & 0.023 & 0.024 $\pm$ 0.001 & 6.10 \\
\bottomrule
\end{tabular}
\caption{Representative heterogeneous unsupervised-jury baseline on pairwise tasks at the largest calibrator-fitting budget. The DS/BT-style jury fits a Dawid--Skene one-coin model \cite{dawid1979maximum} with judge-specific sensitivities and specificities from judge agreement, then orients and isotonic-calibrates the resulting score on the calibration labels; prefix size is selected by validation MSE. The row positions this reliability/jury family relative to the joint-table and scalar selectors used in the regime map.}
\label{tab:jury-baseline}
\end{table}

\begin{table}[t]
\centering
\small
\begin{tabular}{lccccc}
\toprule
Dataset & $n_M$ & MSE sel. MSE & NLL sel. MSE & MSE sel. NLL & NLL sel. NLL \\
\midrule
Arena100K & 50 & 0.237 & 0.237 & 0.697 & 0.676 \\
Arena100K & 100 & 0.235 & 0.235 & 0.697 & 0.670 \\
Arena100K & 200 & 0.233 & 0.232 & 0.702 & 0.670 \\
Arena100K & 400 & 0.232 & 0.232 & 0.665 & 0.664 \\
LLMBar & 20 & 0.221 & 0.221 & 0.704 & 0.632 \\
LLMBar & 50 & 0.211 & 0.210 & 0.647 & 0.621 \\
LLMBar & 100 & 0.197 & 0.198 & 0.635 & 0.647 \\
LLMBar & 200 & 0.191 & 0.191 & 0.607 & 0.578 \\
LLMBar & 300 & 0.183 & 0.184 & 0.557 & 0.559 \\
RewardBench & 50 & 0.022 & 0.020 & 0.123 & 0.094 \\
RewardBench & 100 & 0.022 & 0.022 & 0.139 & 0.109 \\
RewardBench & 200 & 0.022 & 0.022 & 0.115 & 0.108 \\
RewardBench & 400 & 0.022 & 0.021 & 0.106 & 0.095 \\
RewardBench & 800 & 0.023 & 0.022 & 0.101 & 0.096 \\
\bottomrule
\end{tabular}
\caption{Pairwise scalar-selector sensitivity to the validation objective across calibrator-fitting budgets. The candidate menu is unchanged; only the validation criterion changes from MSE to Bernoulli NLL.}
\label{tab:nll-selection-ablation}
\end{table}

\begin{table}[t]
\centering
\scriptsize
\resizebox{\linewidth}{!}{%
\begin{tabular}{lccccc}
\toprule
Dataset & $n_M$ & scalar selector & joint table & tie/error-collapsed table & collapsed $-$ scalar \\
\midrule
RewardBench & 50 & 0.022 $\pm$ 0.001 & 0.026 $\pm$ 0.001 & 0.026 $\pm$ 0.001 & +0.004 \\
 & 100 & 0.022 $\pm$ 0.001 & 0.025 $\pm$ 0.001 & 0.025 $\pm$ 0.001 & +0.003 \\
 & 200 & 0.022 $\pm$ 0.001 & 0.025 $\pm$ 0.001 & 0.025 $\pm$ 0.001 & +0.003 \\
 & 400 & 0.022 $\pm$ 0.001 & 0.025 $\pm$ 0.001 & 0.025 $\pm$ 0.001 & +0.003 \\
 & 800 & 0.023 $\pm$ 0.001 & 0.025 $\pm$ 0.001 & 0.025 $\pm$ 0.001 & +0.002 \\
\addlinespace
LLMBar & 20 & 0.221 $\pm$ 0.004 & 0.226 $\pm$ 0.003 & 0.226 $\pm$ 0.003 & +0.005 \\
 & 50 & 0.211 $\pm$ 0.004 & 0.222 $\pm$ 0.003 & 0.222 $\pm$ 0.003 & +0.011 \\
 & 100 & 0.197 $\pm$ 0.003 & 0.214 $\pm$ 0.003 & 0.214 $\pm$ 0.003 & +0.017 \\
 & 200 & 0.191 $\pm$ 0.004 & 0.214 $\pm$ 0.004 & 0.214 $\pm$ 0.004 & +0.022 \\
 & 300 & 0.183 $\pm$ 0.002 & 0.203 $\pm$ 0.002 & 0.203 $\pm$ 0.002 & +0.021 \\
\addlinespace
Arena100K & 50 & 0.237 $\pm$ 0.002 & 0.239 $\pm$ 0.002 & 0.239 $\pm$ 0.002 & +0.002 \\
 & 100 & 0.235 $\pm$ 0.002 & 0.236 $\pm$ 0.003 & 0.236 $\pm$ 0.003 & +0.001 \\
 & 200 & 0.233 $\pm$ 0.002 & 0.234 $\pm$ 0.002 & 0.234 $\pm$ 0.002 & +0.001 \\
 & 400 & 0.232 $\pm$ 0.002 & 0.233 $\pm$ 0.002 & 0.233 $\pm$ 0.002 & +0.001 \\
\bottomrule
\end{tabular}%
}
\caption{Tie/parse-error encoding check on pairwise tasks across calibrator-fitting budgets. The scalar selector uses the midpoint encoding from the main experiments. The last table column reruns the joint-table validation envelope after merging \texttt{tie} and \texttt{parse\_error} into a shared neutral value, so those categories no longer create separate cells. The positive collapsed--scalar gaps show that the scalar advantage is not explained by giving scalar methods a uniquely favorable neutral encoding.}
\label{tab:tie-parse-fairness}
\end{table}

\begin{table}[t]
\centering
\small
\begin{tabular}{lccccc}
\toprule
Dataset & $n_H$ & 1:1:1 & 1:2:1 & 1:3:1 & 1:1:2 \\
\midrule
RewardBench & 1400 & 0.022 (S) & 0.022 (S) & 0.022 (S) & 0.022 (S) \\
LLMBar & 500 & 0.187 (S) & 0.186 (S) & 0.186 (S) & 0.188 (S) \\
SummEval & 1184 & 0.044 (S) & 0.043 (S) & 0.043 (S) & 0.044 (S) \\
Arena100K & 800 & 0.229 (S) & 0.229 (S) & 0.228 (S) & 0.229 (S) \\
\bottomrule
\end{tabular}
\caption{Fixed deployable-budget allocation sensitivity. For each dataset, the test block is held out and the deployable label budget \(n_H=n_S+n_M+n_V\) is fixed to the largest-budget main-experiment value; columns reallocate that budget across path selection, calibrator fitting, and validation. Each cell reports validation-selected test MSE over a compact menu containing scalar/reliability/stacking candidates and the smoothed joint table on the selection-block information-first path. Parentheses show the modal selected family: S for scalar, T for table.}
\label{tab:fixed-budget-allocation}
\end{table}

\begin{table}[t]
\centering
\small
\begin{tabular}{lccccc}
\toprule
Dataset & scalar & hier. backoff & sel.-heavy & default & val.-heavy \\
\midrule
RewardBench & 0.023 & 0.024 & 0.023 & 0.023 & 0.022 \\
LLMBar & 0.183 & 0.202 & 0.196 & 0.187 & 0.184 \\
SummEval & 0.044 & 0.043 & 0.045 & 0.044 & 0.044 \\
Arena100K & 0.232 & 0.230 & 0.233 & 0.233 & 0.231 \\
\bottomrule
\end{tabular}
\caption{Two additional robustness checks at the largest calibrator-fitting budget. The hierarchical-backoff column validates a prefix-smoothed table on the information-first path, selecting \(K\) and shrinkage by validation MSE. The last three columns rerun the scalar-family selector while shifting labels between the selection and validation blocks and keeping the calibration and test sizes fixed.}
\label{tab:backoff-split-sensitivity}
\end{table}

\begin{table}[t]
\centering
\small
\begin{tabular}{lcccc}
\toprule
Dataset & additive risk & pairwise risk & pairwise-additive & pairwise selected \\
\midrule
Arena100K & 0.2278 & 0.2291 & 0.0013 & 0.10 \\
LLMBar & 0.1809 & 0.1832 & 0.0024 & 0.30 \\
RewardBench & 0.0214 & 0.0214 & 0.0000 & 0.30 \\
SummEval & 0.0431 & 0.0435 & 0.0004 & 0.33 \\
\bottomrule
\end{tabular}
\caption{Real-data interaction diagnostic for scalar aggregation at the largest calibration budget. The additive column is the best all-seven ridge/logistic stacker; the pairwise column is the best all-seven stacker with second-order products. The last column reports the share of validation-selected scalar aggregators that use pairwise interaction features.}
\label{tab:interaction-diagnostic}
\end{table}

\begin{table}[t]
\centering
\small
\begin{tabular}{lcc}
\toprule
Diagnostic & Estimate & Uncertainty / test \\
\midrule
Hard rule agreement over dataset-budget cells & 16/20 & one-sided binomial $p=0.006$ \\
Split-level proxy AUC for scalar winner & 0.573 & cluster bootstrap 95\% CI [0.522, 0.624] \\
Within-cluster permutation test for AUC & -- & $p=0.002$ \\
Scalar-minus-table proxy-score gap & 0.476 & cluster bootstrap 95\% CI [0.221, 0.754] \\
\bottomrule
\end{tabular}
\caption{Statistical check for the theory-proxy winner diagnostic. The hard rule gives a coarse aggregate check. The split-level tests use the continuous proxy score \(z(H_K/n_M)+z(\mathrm{unseen})+z(\mathrm{pairwise\;gain})\) to predict whether validation-selected scalar aggregation beats validation-selected table calibration on held-out test risk. Bootstrap resamples dataset-budget clusters; the permutation test shuffles winners within each dataset-budget cluster.}
\label{tab:theory-proxy-winner}
\end{table}

\begin{table}[t]
\centering
\small
\begin{tabular}{lccc}
\toprule
Dataset & global fallback & lower-prefix fallback & best $\alpha$/fallback \\
\midrule
RewardBench & 0.024 $\pm$ 0.001 & 0.024 $\pm$ 0.001 & $\alpha=2.0$, lower-prefix: 0.023 \\
LLMBar & 0.204 $\pm$ 0.002 & 0.205 $\pm$ 0.002 & $\alpha=2.0$, lower-prefix: 0.203 \\
SummEval & 0.044 $\pm$ 0.001 & 0.044 $\pm$ 0.001 & $\alpha=2.0$, global: 0.044 \\
Arena100K & 0.231 $\pm$ 0.002 & 0.231 $\pm$ 0.002 & $\alpha=2.0$, lower-prefix: 0.230 \\
\bottomrule
\end{tabular}
\caption{Joint-table sensitivity at the largest calibrator-fitting budget, using an information-first path and validation-selected prefix size. The table compares the default global-mean fallback against a lower-prefix fallback and reports the best alpha/fallback setting from the sweep. For SummEval, this sensitivity is after the fixed five-bin scalar judge discretization described in the appendix.}
\label{tab:table-sensitivity}
\end{table}

\begin{table}[!htbp]
\centering
\small
\begin{tabular}{lrrrr}
\toprule
Dataset & table & scalar & residual-shrinkage & selected base \\
\midrule
RewardBench & 0.0251 & 0.0227 & 0.0223 & one\_coin\_iso \\
LLMBar & 0.2031 & 0.1826 & 0.1833 & logistic \\
SummEval & 0.0454 & 0.0435 & 0.0437 & ridge\_iso \\
Arena100K & 0.2331 & 0.2321 & 0.2324 & one\_coin\_iso \\
\bottomrule
\end{tabular}
\caption{Prototype residual-shrinkage hybrid at each dataset's largest calibrator-fitting budget. The method first fits a scalar baseline, then adds a count-shrunk residual table on joint output patterns and selects base family, prefix size, and shrinkage strength by validation risk. This diagnostic reports an intermediate family between scalar stackers and full joint tables.}
\label{tab:residual-shrinkage-hybrid}
\end{table}

\FloatBarrier

\subsection{Calibration-Frontier Path Diagnostic}

We evaluated a calibration-frontier path search (CFPS) generator that ranks judge subsets by
selection-block oracle gain and exact binomial estimates of the cell-pressure term in
Eq.~\eqref{eq:cell-pressure}, keeps a capped Pareto frontier
(\(\texttt{cfps\_max\_paths}=32\)), and then applies the same validation selector.
Table~\ref{tab:cfps-path} compares this generator against the original split-wise
validation envelope over accuracy, gain-complexity, and random paths. The result is a
calibration-frontier view of path generation: CFPS wins some low-budget cells, while
the information-first path remains competitive across datasets. The essential
decision is validation over prefix size and aggregation family under finite
calibration labels.

\begin{table}[!htbp]
\centering
\small
\begin{tabular}{lrrrrrr}
\toprule
Dataset & $n_M$ & baseline & CFPS & +CFPS & $\Delta_{+}$ & CFPS sel. \\
\midrule
Arena100K & 400 & 0.2331 & 0.2343 & 0.2341 & +0.0010 & 0.77 \\
LLMBar & 300 & 0.2031 & 0.2043 & 0.2049 & +0.0019 & 0.73 \\
RewardBench & 800 & 0.0251 & 0.0253 & 0.0260 & +0.0009 & 0.67 \\
SummEval & 800 & 0.0454 & 0.0461 & 0.0461 & +0.0007 & 0.73 \\
\bottomrule
\end{tabular}
\caption{Calibration-frontier path search (CFPS) at each dataset's largest calibrator-fitting budget. The baseline is the split-wise validation envelope over accuracy, gain-complexity, and random paths; +CFPS adds the calibration-frontier candidates to the same validation envelope. Negative $\Delta_{+}$ means the augmented menu has lower test risk.}
\label{tab:cfps-path}
\end{table}

\FloatBarrier

\subsection{Cold-Start Few-Label Diagnostic}

We also simulate a deployment setting in which the dataset itself is a new target task
and task type provides the prior information. For each real benchmark we expose
\(n_h\in\{8,16,32,64\}\) grouped target labels for selection and calibration, then
evaluate on a held-out grouped test block. The task-type prior uses one-coin
reliability for pairwise tasks and ridge calibration for scalar tasks. Unguarded
\fcps{} always deploys the cross-validated selector over path, prefix size, and aggregator;
guarded \fcps{} deviates from the prior when the selected candidate's
cross-validated upper one-SE risk is below the prior's lower one-SE risk. Table~\ref{tab:cold-start-few-label}
shows that this extreme low-label setting is prior-dominated: the guard stabilizes
selection and adapts when the validation interval gives a clear improvement
over the task-type prior. We therefore treat cold-start selection as a separate
deployment problem with its own prior and uncertainty requirements.

\begin{table}[!htbp]
\centering
\scriptsize
\setlength{\tabcolsep}{3pt}
\begin{tabular}{lrrrrrl}
\toprule
Dataset & $n_h$ & prior & unguarded FCPS & guarded FCPS & $\Delta_g$ & guarded family \\
\midrule
Arena100K & 16 & 0.2670 & 0.2671 & 0.2686 & +0.0016 & One-coin reliability + isotonic (0.42) \\
Arena100K & 64 & 0.2399 & 0.2467 & 0.2427 & +0.0028 & One-coin reliability + isotonic (0.25) \\
LLMBar & 16 & 0.2600 & 0.2541 & 0.2560 & -0.0040 & One-coin reliability + isotonic (0.58) \\
LLMBar & 64 & 0.2053 & 0.2135 & 0.2051 & -0.0001 & One-coin reliability + isotonic (0.08) \\
RewardBench & 16 & 0.0340 & 0.0406 & 0.0340 & +0.0000 & One-coin reliability + isotonic (0.00) \\
RewardBench & 64 & 0.0251 & 0.0323 & 0.0275 & +0.0024 & One-coin reliability + isotonic (0.17) \\
SummEval & 16 & 0.0664 & 0.0781 & 0.0664 & +0.0000 & Ridge + isotonic (0.00) \\
SummEval & 64 & 0.0513 & 0.0530 & 0.0518 & +0.0005 & Ridge + isotonic (0.17) \\
\bottomrule
\end{tabular}
\caption{Cold-start deployment with few human calibration labels over 12 random grouped splits. Each target dataset is treated as a new task; only $n_h$ target labels are used for selection and calibration, then the selected predictor is refit on all $n_h$ labels and evaluated on a held-out grouped test block. The prior baseline uses only task type: one-coin reliability for pairwise tasks and ridge calibration for scalar tasks. Unguarded FCPS always deploys the cross-validated selector. Guarded FCPS deviates from the task-type prior only when the selected candidate's cross-validated upper one-SE risk is below the prior's cross-validated lower one-SE risk. Negative $\Delta_g$ favors guarded FCPS; the parenthetical entry is the adaptation rate.}
\label{tab:cold-start-few-label}
\end{table}

\FloatBarrier

\subsection{External Safety, Math, and Code Binary Validation}
\label{app:external-binary-validation}

We also instantiate the theory as an out-of-domain prediction exercise on three
binary targets: JailbreakBench safety, GSM8K math correctness, and MBPP code
correctness. Table~\ref{tab:external-binary-validation-protocol} shows the common
format. The prediction is made before treating the held-out results as evidence:
increasing \(K\) should improve the oracle information term when the additional
judge outputs expose useful structure, while the joint-pattern table should also
show larger support and unseen-pattern pressure. Therefore finite validation should
often stop at an intermediate prefix.

\begin{table}[!htbp]
\centering
\small
\begin{tabular}{llll}
\toprule
Task & Source dataset & Positive label & Judge alphabet \\
\midrule
Safety & JailbreakBench judge comparison & safe/refusal response & safe, unsafe, uncertain \\
Math & GSM8K correctness & correct final answer & correct, incorrect, uncertain \\
Code & MBPP correctness & code passes task tests & correct, incorrect, uncertain \\
\bottomrule
\end{tabular}
\caption{External binary validation tasks instantiated in the same seven-judge
finite-calibration format. Each task produces a labeled JSONL panel with grouped
source identifiers, seven deterministic judge outputs, and the same
selection/calibration/validation/test protocol.}
\label{tab:external-binary-validation-protocol}
\end{table}

The completed runs match the main qualitative prediction: individual judges can be
strong, but the validation-selected configurations stop at intermediate average
prefix sizes and the full seven-judge table has higher support and unseen-pattern
pressure. Table~\ref{tab:external-binary-validation-results} summarizes the three
external runs. Math is near-ceiling in this first objective construction, so its
main value is as a complexity-growth sanity check; safety and code give more
informative finite-selection stress tests. Table~\ref{tab:external-binary-theory-validation}
gives the prediction-to-observation map across all three external tasks.

\begin{table}[!htbp]
\centering
\small
\begin{tabular}{lcccccc}
\toprule
Task & Best single acc. & Best strategy & \(\ncal\) & selected \(K\) & test risk & unseen, \(K{=}1\to7\) \\
\midrule
Safety & 0.850 & gain-complexity & 96 & 4.50 & 0.115 & 0.000 \(\to\) 0.210 \\
Math & 1.000 & gain-complexity & 64 & 2.17 & 0.004 & 0.000 \(\to\) 0.033 \\
Code & 0.980 & gain-complexity & 96 & 2.63 & 0.011 & 0.003 \(\to\) 0.053 \\
\bottomrule
\end{tabular}
\caption{Completed external binary validation runs. Each task uses 300 labeled
rows, the same seven-judge pool, grouped splits by source item, and the
joint-table prefix selector. The unseen column reports the mean held-out unseen
pattern rate from \(K=1\) to \(K=7\) on the gain-complexity path at
\(\ncal=96\), except math's best row occurs at \(\ncal=64\); its unseen trend is
shown at \(\ncal=96\) for comparability. Math is near-ceiling in this objective
construction, so this task primarily contributes a clean complexity-growth check.}
\label{tab:external-binary-validation-results}
\end{table}

\begin{table}[!htbp]
\centering
\scriptsize
\setlength{\tabcolsep}{3pt}
\begin{tabular}{p{0.24\linewidth}p{0.21\linewidth}p{0.21\linewidth}p{0.21\linewidth}}
\toprule
Theory prediction & Safety & Math & Code \\
\midrule
Strong judges do not automatically determine the deployed panel &
Best single accuracy is 0.850; validation selects \(K=4.50\). &
Best single accuracy is 1.000; validation still selects \(K=2.17\), but this task is near-ceiling. &
Best single accuracy is 0.980; validation selects \(K=2.63\). \\
Pattern complexity grows with panel size &
\(Q_K: 2.3\to384\); entropy support \(2.11\to14.02\). &
\(Q_K: 2.0\to128\); entropy support \(2.00\to3.94\). &
\(Q_K: 2.4\to192\); entropy support \(2.01\to5.04\). \\
Sparse calibration creates unseen-pattern pressure &
Unseen rate \(0.000\to0.210\). &
Unseen rate \(0.000\to0.033\). &
Unseen rate \(0.003\to0.053\). \\
Finite risk favors intermediate prefixes &
Best row: gain-complexity, \(\ncal=96\), \(K=4.50\), risk 0.115. &
Best row: gain-complexity, \(\ncal=64\), \(K=2.17\), risk 0.004. &
Best row: gain-complexity, \(\ncal=96\), \(K=2.63\), risk 0.011. \\
\bottomrule
\end{tabular}
\caption{Prediction-to-observation map for the external binary validation tasks.
All three use the same seven-judge pool and grouped finite-calibration protocol.
The reported complexity and unseen trends compare \(K=1\) to \(K=7\) on the
gain-complexity path at \(\ncal=96\). The math task is near-ceiling in this first
construction, so it is interpreted mainly as a complexity-growth check.}
\label{tab:external-binary-theory-validation}
\end{table}

\FloatBarrier

\subsection{New Calibration-Growth Dataset}

To isolate the effect of increasing calibration labels, we generated a fresh JSONL
panel dataset with seven binary judges and three regimes: additive, mixed, and
six-way parity. We then increased the calibrator-fitting budget from 16 to 2048 labels and
compared validation-selected prefix sizes within a joint-table family, an additive
ridge family, and a pairwise-ridge family. Table~\ref{tab:calibration-growth-new-dataset}
reports the best scalar risk minus the joint-table risk. The additive regime remains
scalar-favored at moderate and large budgets. In contrast, the mixed and parity
regimes favor the table: for the parity regime, the selected table uses \(K=6\), its
held-out unseen-pattern rate falls from \(0.768\) at \(\ncal=16\) to \(0\) by
\(\ncal=512\), and its test MSE falls from \(0.224\) to \(0.061\) at that budget. The
\(+0.053\) table entry at \(\ncal=1024\) is the scalar-minus-table gap, not the table
risk. This is the clean positive counterpart to the real-benchmark results: once the
target truly contains a high-order interaction, additional calibration labels can
make the nonparametric aggregator worthwhile.

\begin{table}[!htbp]
\centering
\small
\begin{tabular}{lrrrrr}
\toprule
Regime & 16 & 64 & 256 & 1024 & 2048 \\
\midrule
Additive & -0.007 & -0.003 & -0.004 & -0.002 & -0.001 \\
Mixed & \textbf{+0.014} & \textbf{+0.014} & \textbf{+0.043} & \textbf{+0.018} & \textbf{+0.010} \\
Parity & \textbf{+0.046} & \textbf{+0.101} & \textbf{+0.123} & \textbf{+0.053} & \textbf{+0.010} \\
\bottomrule
\end{tabular}
\caption{New calibration-growth dataset. Entries are mean test-MSE gaps after validation-selecting prefix size within each family: best scalar aggregator minus joint table. Positive values favor the joint table; bold marks gaps larger than 0.001. Thus the parity \(0.053\) entry at \(n_M=1024\) is a gap, not the table risk. The additive regime remains scalar-favored, while the high-order parity regime tests when enough calibration labels are available for a nonparametric table to exploit interactions.}
\label{tab:calibration-growth-new-dataset}
\end{table}

\FloatBarrier

\subsection{Semi-Real Pattern Calibration-Growth Dataset}

To make the interaction stress test less stylized, we also construct a semi-real
dataset by resampling seven-judge output patterns from the real RewardBench, LLMBar,
and Arena100K panels, then assigning controlled additive, mixed, and parity targets to
those real patterns. This preserves the empirical judge correlations and pattern
sparsity while letting us know whether the target contains an interaction. Table~\ref{tab:semireal-calibration-growth}
shows the expected boundary: additive targets become scalar-favored after the
smallest budgets, while mixed and parity targets favor the joint table. In the parity
regime, the selected table keeps \(K\approx6\), unseen rate falls from \(0.273\) at
\(\ncal=16\) to \(0.007\) at \(\ncal=2048\), and the table advantage remains positive.

\begin{table}[!htbp]
\centering
\small
\begin{tabular}{lrrrrr}
\toprule
Regime & 16 & 64 & 256 & 1024 & 2048 \\
\midrule
Additive & \textbf{+0.003} & +0.000 & -0.001 & -0.002 & -0.001 \\
Mixed & \textbf{+0.010} & \textbf{+0.005} & \textbf{+0.011} & \textbf{+0.006} & \textbf{+0.004} \\
Parity & \textbf{+0.065} & \textbf{+0.055} & \textbf{+0.030} & \textbf{+0.015} & \textbf{+0.009} \\
\bottomrule
\end{tabular}
\caption{Semi-real pattern calibration-growth dataset. Entries are mean test-MSE gaps after validation-selecting prefix size within each family: best scalar aggregator minus joint table. Positive values favor the joint table; bold marks gaps larger than 0.001. The judge-output patterns are resampled from real RewardBench, LLMBar, and Arena100K seven-judge outputs, while the target rule is controlled. Additive targets become scalar-favored after the smallest budgets; mixed and parity targets favor the joint table.}
\label{tab:semireal-calibration-growth}
\end{table}

\begin{table}[!htbp]
\centering
\small
\begin{tabular}{llccc}
\toprule
Regime & Budget & Joint table & Ridge + isotonic & Logistic \\
\midrule
Additive & 50 & 0.187 & 0.178 & \textbf{0.171} \\
Additive & 3000 & 0.159 & 0.158 & \textbf{0.157} \\
XOR interaction & 50 & \textbf{0.140} & 0.219 & 0.251 \\
XOR interaction & 3000 & \textbf{0.117} & 0.203 & 0.236 \\
AND interaction & 50 & 0.139 & \textbf{0.137} & 0.156 \\
AND interaction & 3000 & \textbf{0.107} & 0.115 & 0.129 \\
\bottomrule
\end{tabular}
\caption{Controlled aggregator-regime experiment. Each entry is test MSE after selecting prefix size by validation within an aggregator family; bold marks the best family for that regime and budget. Additive signals favor low-dimensional stacking, while interaction signals favor the joint table once the calibrator-fitting budget is large enough.}
\label{tab:controlled-aggregator-regimes}
\end{table}

\begin{table}[t]
\centering
\small
\begin{tabular}{lrrrr}
\toprule
Interaction strength $\gamma$ & $n_M=50$ & $n_M=200$ & $n_M=1000$ & $n_M=3000$ \\
\midrule
0.0 & -0.024 & -0.013 & -0.003 & +0.000 \\
0.3 & -0.020 & \textbf{+0.008} & \textbf{+0.014} & \textbf{+0.012} \\
0.5 & \textbf{+0.046} & \textbf{+0.053} & \textbf{+0.056} & \textbf{+0.056} \\
0.7 & \textbf{+0.110} & \textbf{+0.117} & \textbf{+0.117} & \textbf{+0.135} \\
1.0 & \textbf{+0.171} & \textbf{+0.186} & \textbf{+0.182} & \textbf{+0.188} \\
\bottomrule
\end{tabular}
\caption{Interaction-strength sweep. Entries are mean test-MSE gaps after validation selection within each family: best scalar aggregator minus joint table. Positive values mean the joint table has lower risk; values above 0.001 are bolded. As interaction strength and calibrator-fitting budget grow, the winner shifts from low-dimensional scalar aggregation toward the joint table.}
\label{tab:interaction-strength-sweep}
\end{table}

\begin{table}[t]
\centering
\small
\begin{tabular}{lccc}
\toprule
Diagnostic & Estimate & SE & controlled units \\
\midrule
$H_K/n_M \rightarrow$ test risk & 0.178 & 0.009 & 12600 residualized prefixes \\
unseen rate $\rightarrow$ test risk & 0.199 & 0.009 & 12600 residualized prefixes \\
\bottomrule
\end{tabular}
\caption{Controlled complexity diagnostics. Predictors and test risk are demeaned within dataset--budget--path groups before fitting a standardized one-variable regression. Positive estimates indicate that, even within the same dataset, calibrator-fitting budget, and path rule, larger complexity pressure or unseen pattern rate is associated with higher test risk. This is diagnostic evidence, not a causal estimate.}
\label{tab:controlled-complexity}
\end{table}

\begin{table}[t]
\centering
\small
\begin{tabular}{lccc}
\toprule
Dataset & groups & mean within-group corr. & positive fraction \\
\midrule
RewardBench & 15 & 0.037 & 0.53 \\
LLMBar & 15 & 0.127 & 0.67 \\
SummEval & 18 & 0.484 & 1.00 \\
Arena100K & 12 & 0.331 & 1.00 \\
\bottomrule
\end{tabular}
\caption{Within-dataset, within-budget, within-path correlations between $H_K/n_M$ and test risk. Each group varies over splits and prefix sizes while holding the major confounders fixed.}
\label{tab:within-group-correlation}
\end{table}

\FloatBarrier

\begin{figure}[t]
  \centering
  \includegraphics[width=0.98\linewidth]{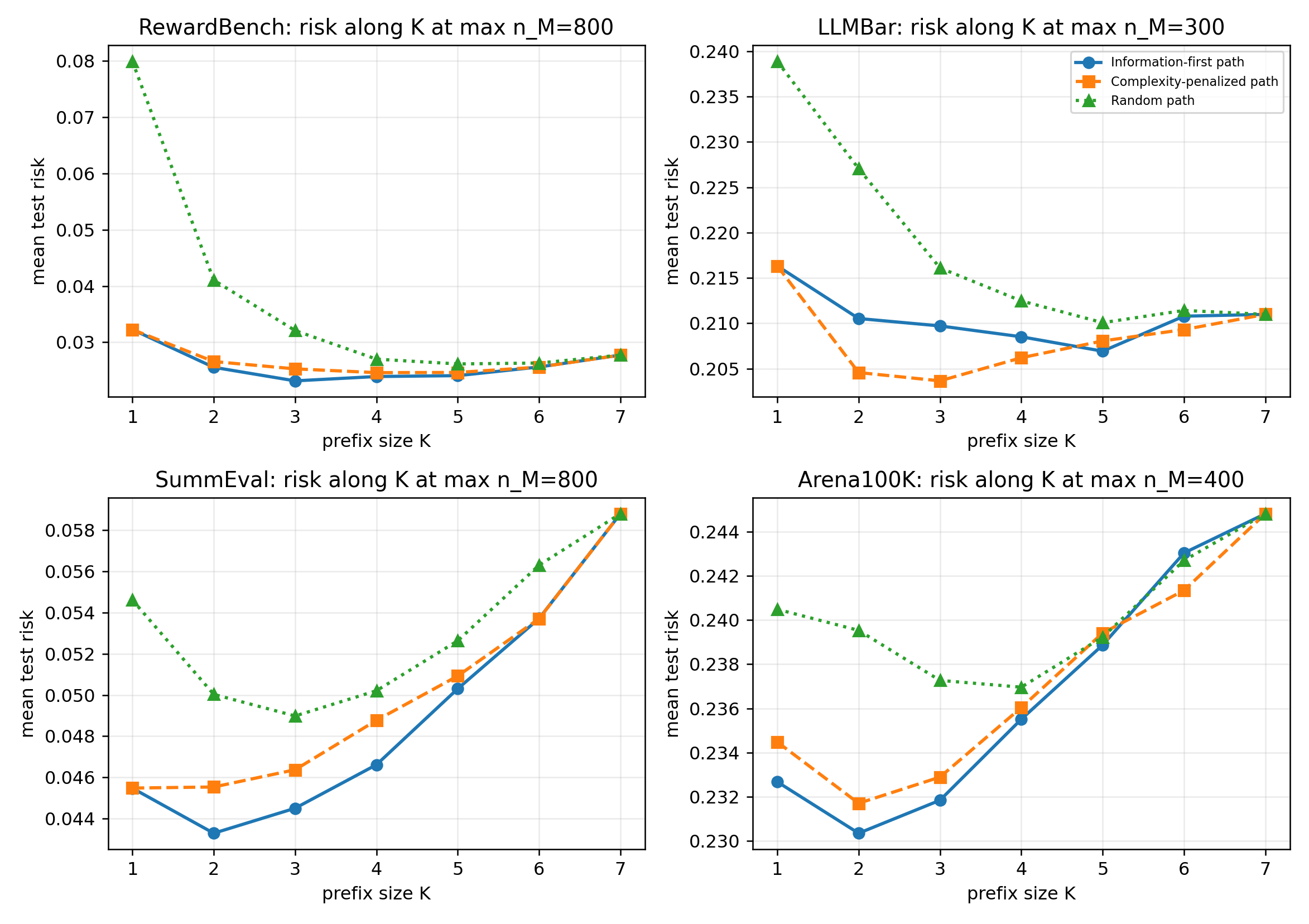}
  \caption{Joint-table finite-risk curves over prefix size \(K\) at the largest
  available calibrator-fitting budget for each dataset.}
  \label{fig:risk-k}
\end{figure}

\begin{figure}[t]
  \centering
  \includegraphics[width=0.98\linewidth]{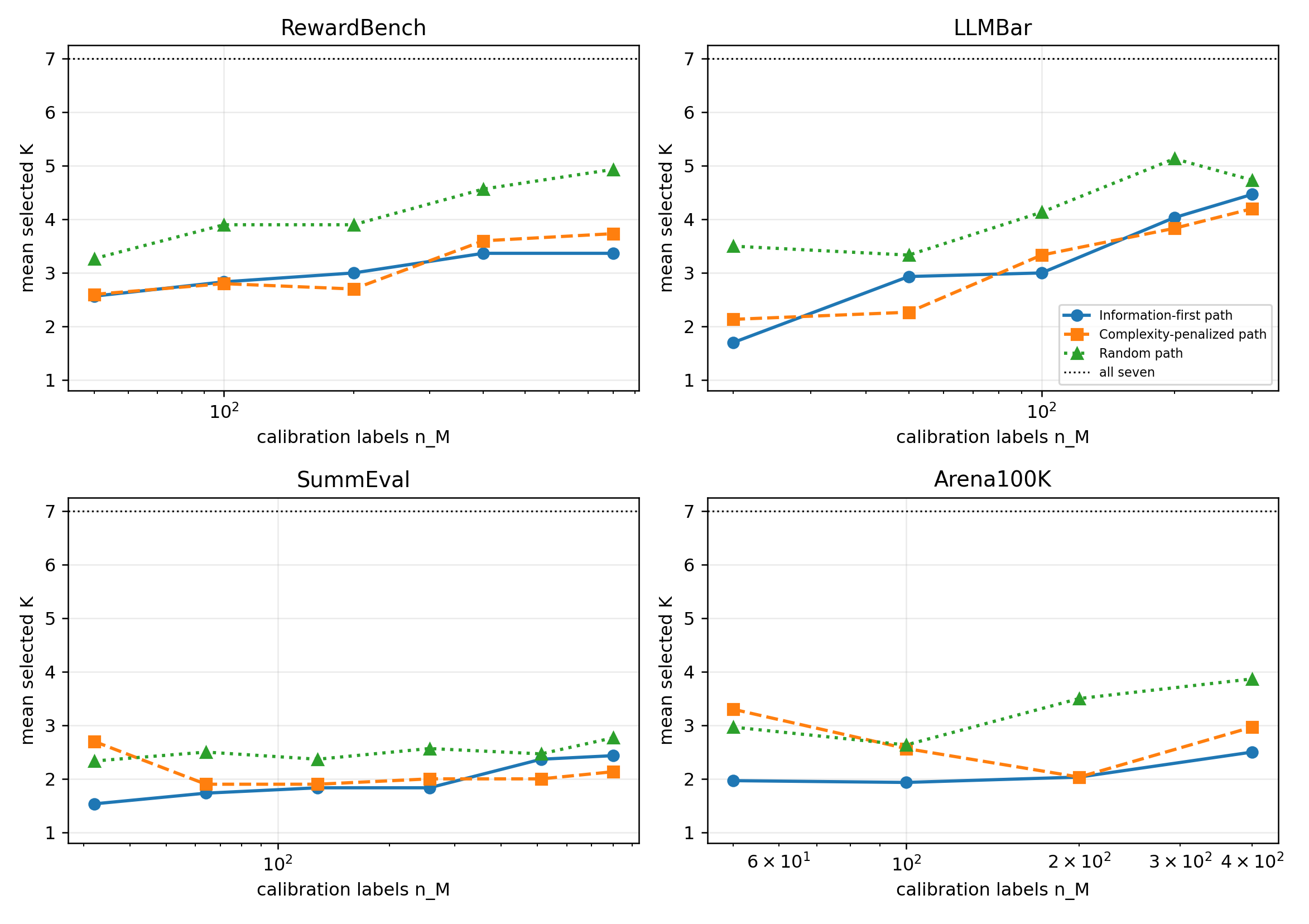}
  \caption{Selected joint-table prefix size as a function of calibrator-fitting budget.}
  \label{fig:selected-k}
\end{figure}

\begin{figure}[t]
  \centering
  \includegraphics[width=0.98\linewidth]{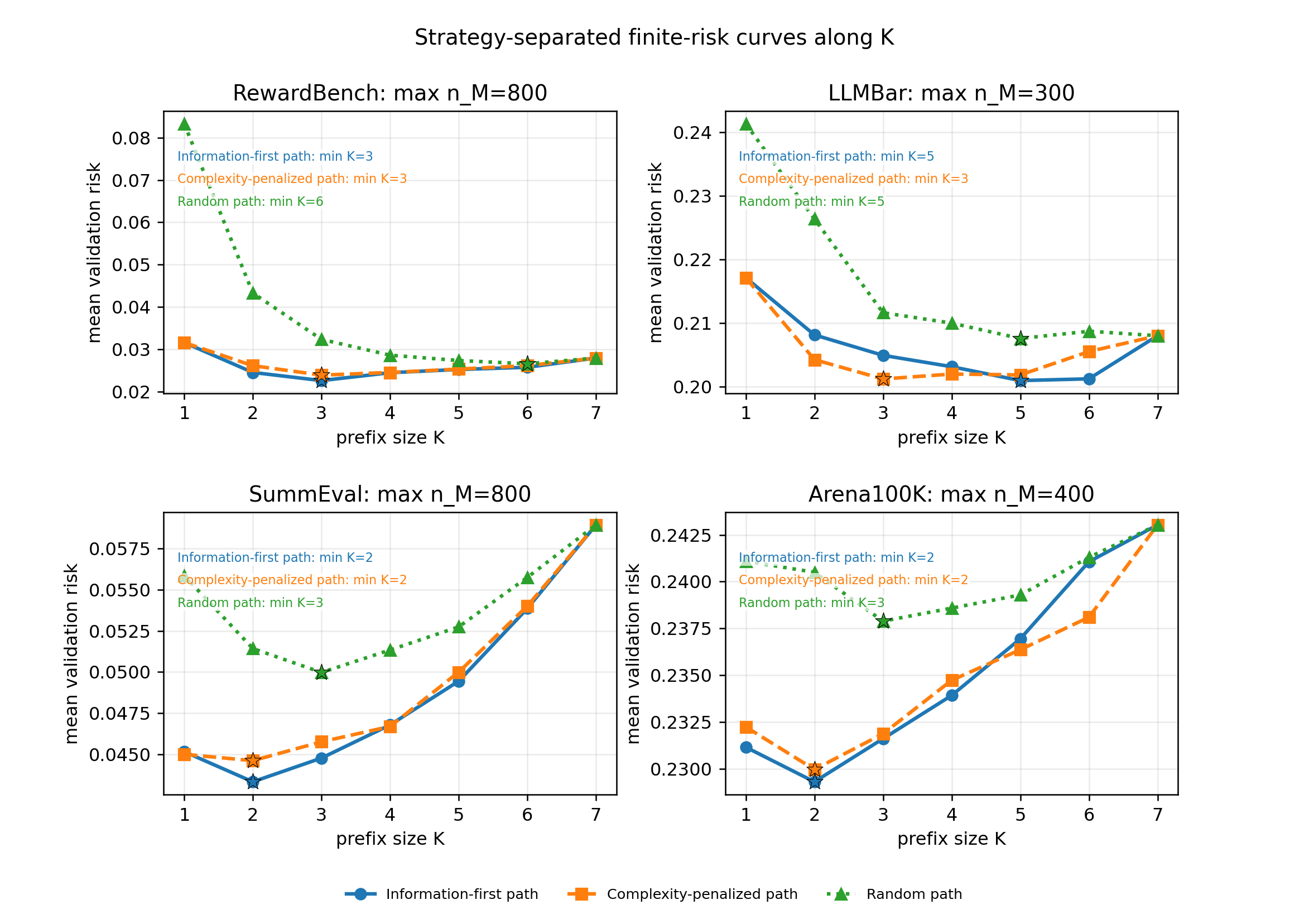}
  \caption{Strategy-separated joint-table validation-risk curves over prefix size
  \(K\). Stars mark each path's mean validation-risk minimum.}
  \label{fig:validation-risk}
\end{figure}

\begin{figure}[t]
  \centering
  \includegraphics[width=0.98\linewidth]{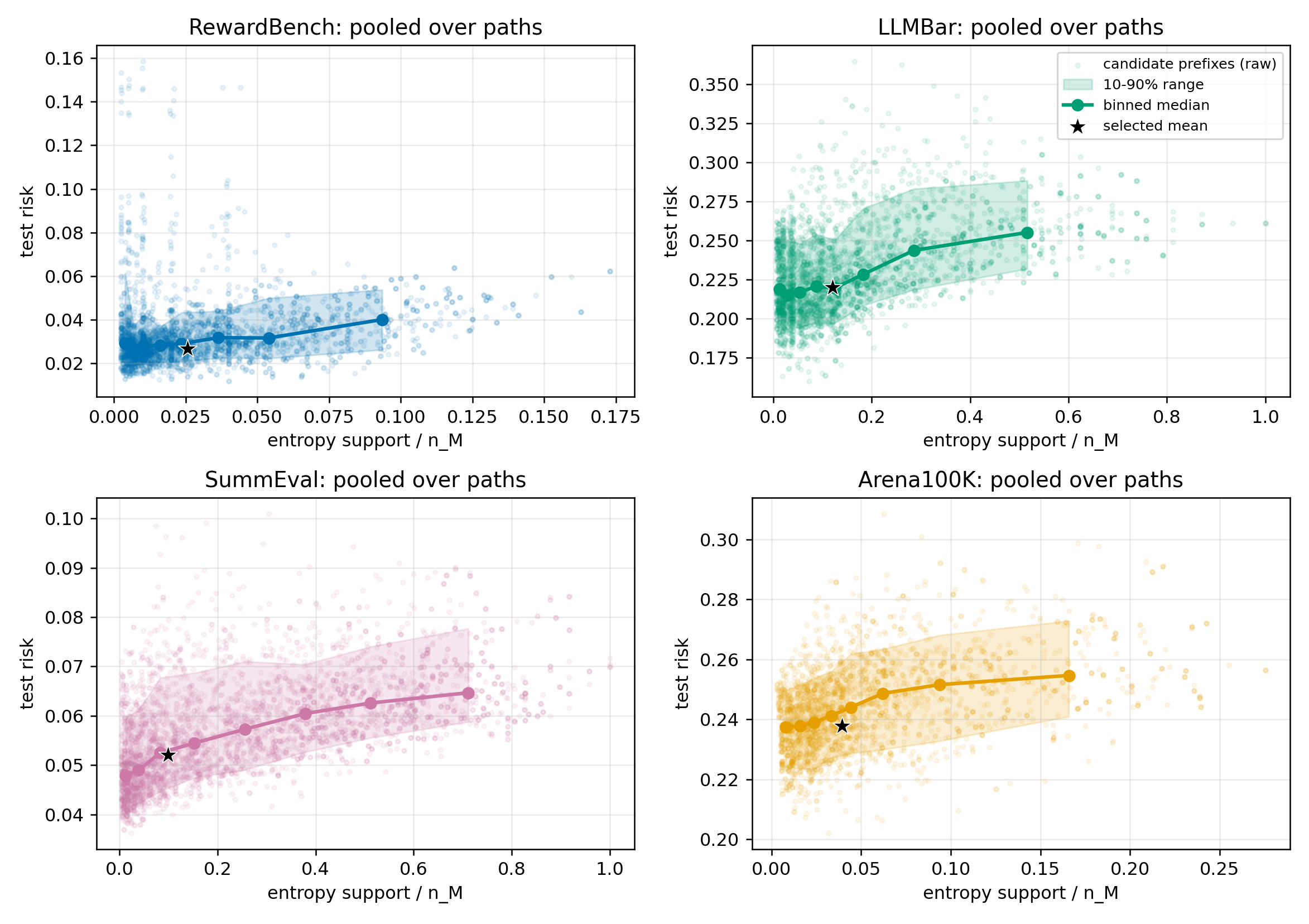}
  \caption{Joint-table finite risk versus the complexity proxy \(H_K/\ncal\), pooled
  over the three path rules.}
  \label{fig:complexity-risk}
\end{figure}

\begin{figure}[t]
  \centering
  \includegraphics[width=0.98\linewidth]{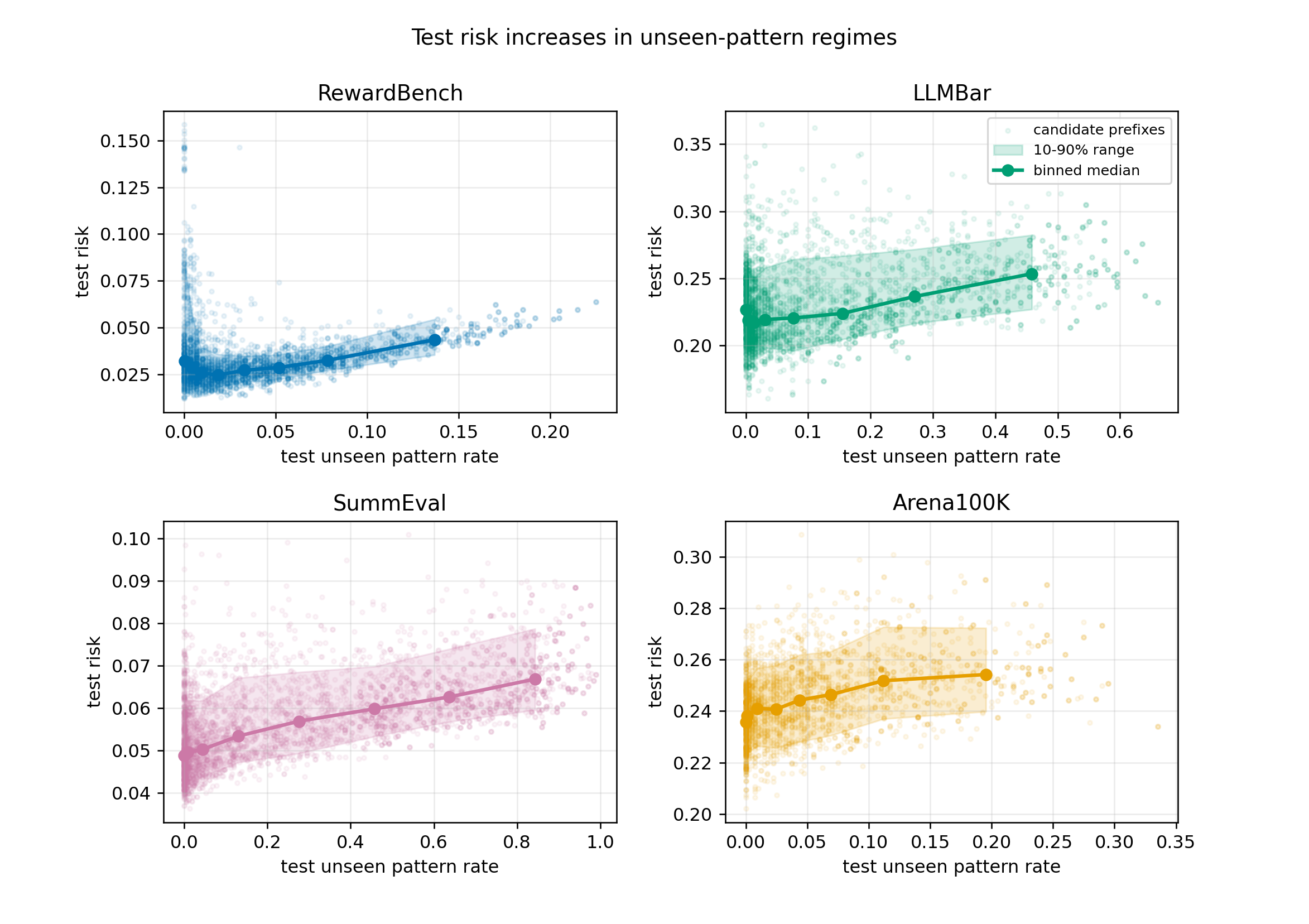}
  \caption{Joint-table test risk as a function of test-time unseen pattern rate.}
  \label{fig:unseen-risk}
\end{figure}

\begin{figure}[t]
  \centering
  \includegraphics[width=0.98\linewidth]{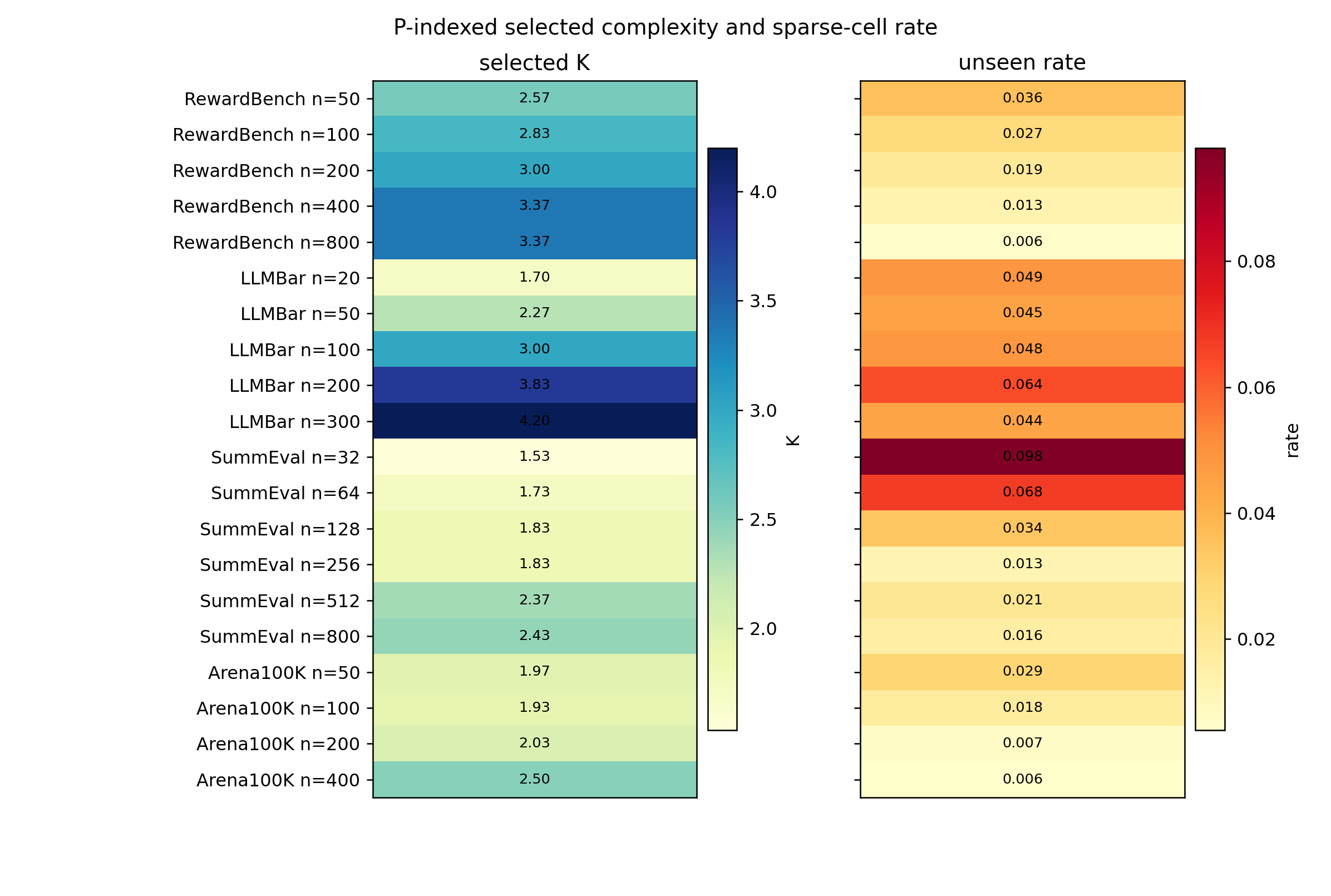}
  \caption{Distribution-indexed selected complexity and sparse-cell rate.}
  \label{fig:distribution-summary}
\end{figure}

\section{Additional Diagnostics}

\subsection{Path-Separated Complexity Diagnostic}

\begin{figure}[t]
  \centering
  \includegraphics[width=0.98\linewidth]{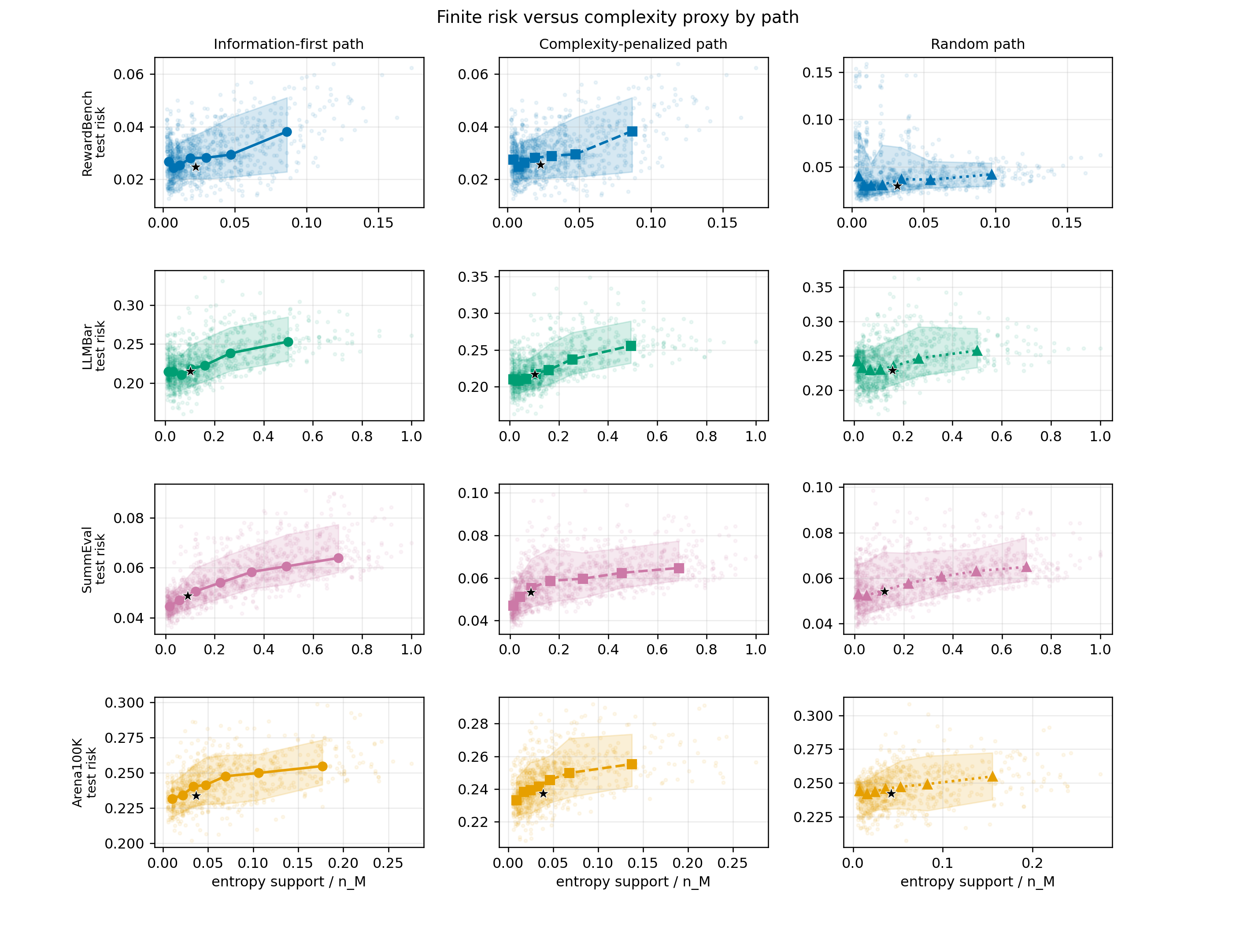}
  \caption{Path-separated version of Figure~\ref{fig:complexity-risk}. Each row is a
  dataset and each column is a path rule. The plot separates information-first,
  complexity-penalized, and random paths so the complexity-pressure pattern can be
  read within each path family.}
  \label{fig:complexity-risk-path}
\end{figure}

Figure~\ref{fig:complexity-risk-path} repeats the complexity-pressure diagnostic
separately for each path rule. Most cells show that higher effective support per
calibration label corresponds to higher median risk or wider risk tails.
Path-specific risk-complexity behavior differs by dataset, and the information-first
path remains a strong practical baseline.

\subsection{Marginal Stopping Diagnostic}

\begin{figure}[t]
  \centering
  \includegraphics[width=0.98\linewidth]{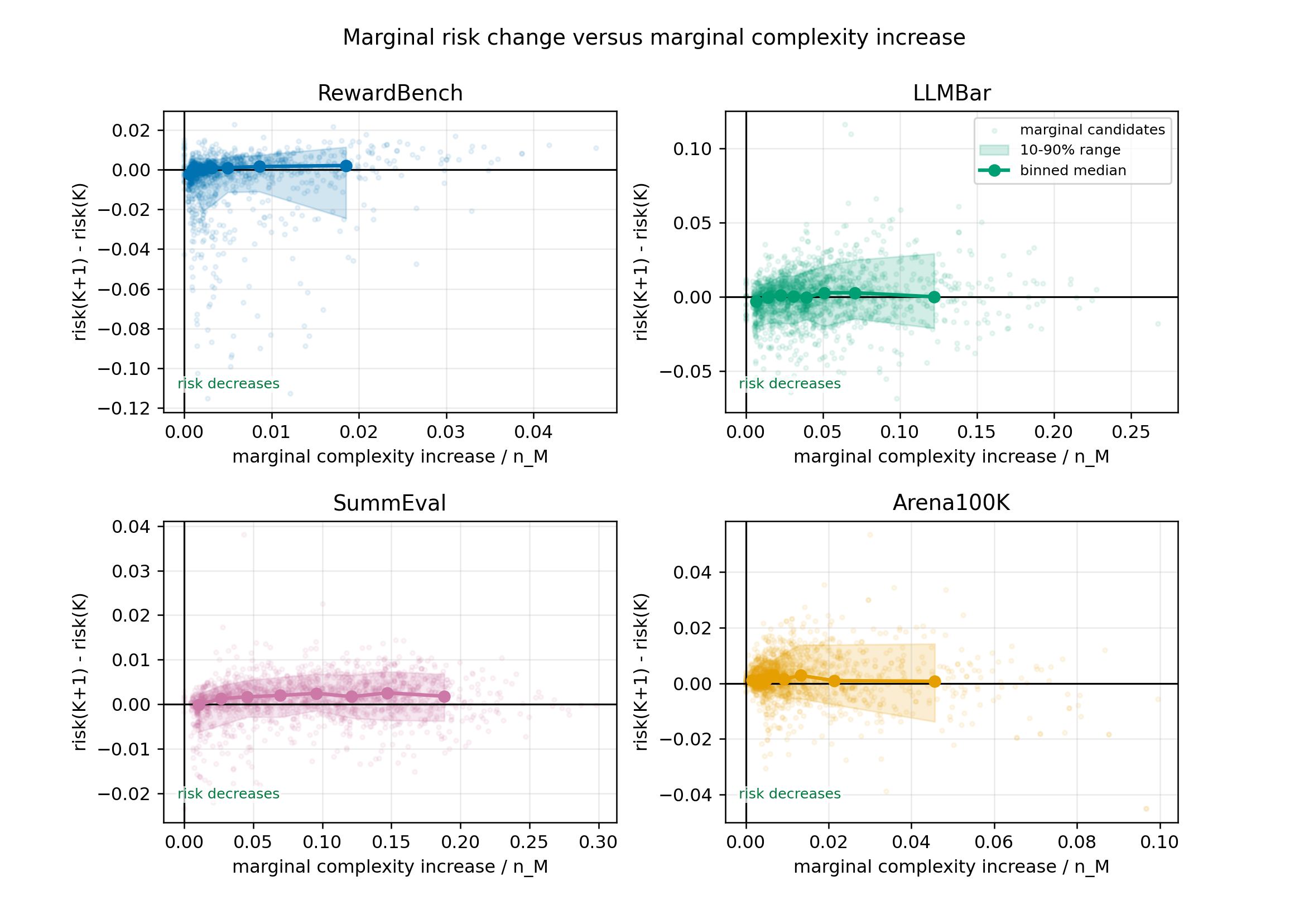}
\caption{Marginal risk change versus marginal complexity increase when moving from
  \(K\) to \(K+1\). Values below zero indicate that adding the next judge reduces risk.
  The binned median and 10--90\% band show that larger marginal complexity increases
  have weaker and more variable risk reductions.}
  \label{fig:marginal}
\end{figure}

Equation~\eqref{eq:finite-risk} suggests a marginal stopping principle. A new judge
should be added when its marginal oracle-information gain exceeds its
calibration-complexity cost. Figure~\ref{fig:marginal} visualizes this principle by
plotting the change in test risk from \(K\) to \(K+1\) against the corresponding
increase in complexity pressure. Points below zero are beneficial additions. The
median trend is close to zero and the tails widen as marginal complexity increases,
indicating that extra judges have increasingly variable finite-risk value.

\subsection{Uncertainty Diagnostics}

\begin{figure}[t]
  \centering
  \begin{subfigure}{0.49\linewidth}
    \centering
    \includegraphics[width=\linewidth]{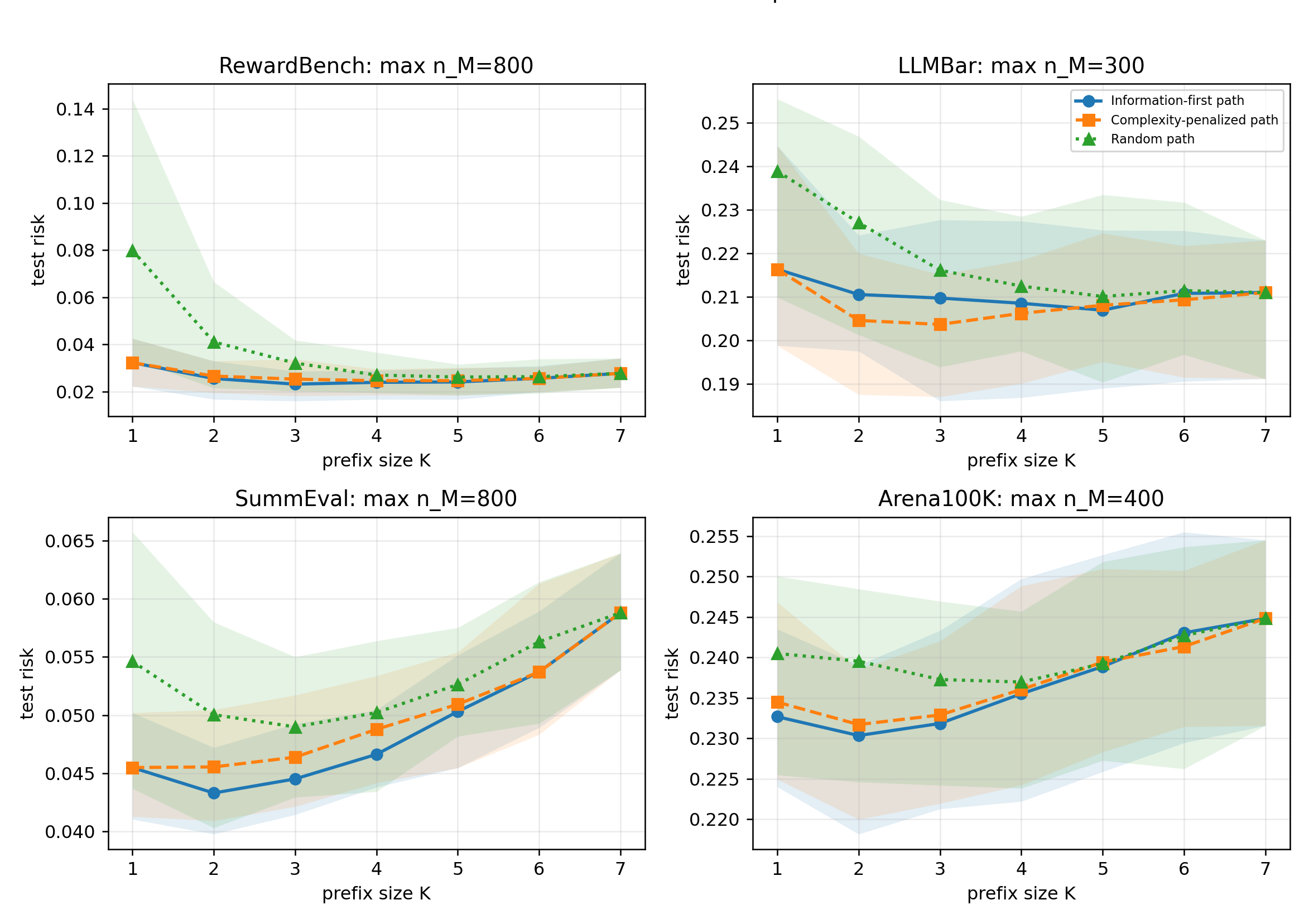}
    \caption{Risk envelope.}
  \end{subfigure}
  \hfill
  \begin{subfigure}{0.49\linewidth}
    \centering
    \includegraphics[width=\linewidth]{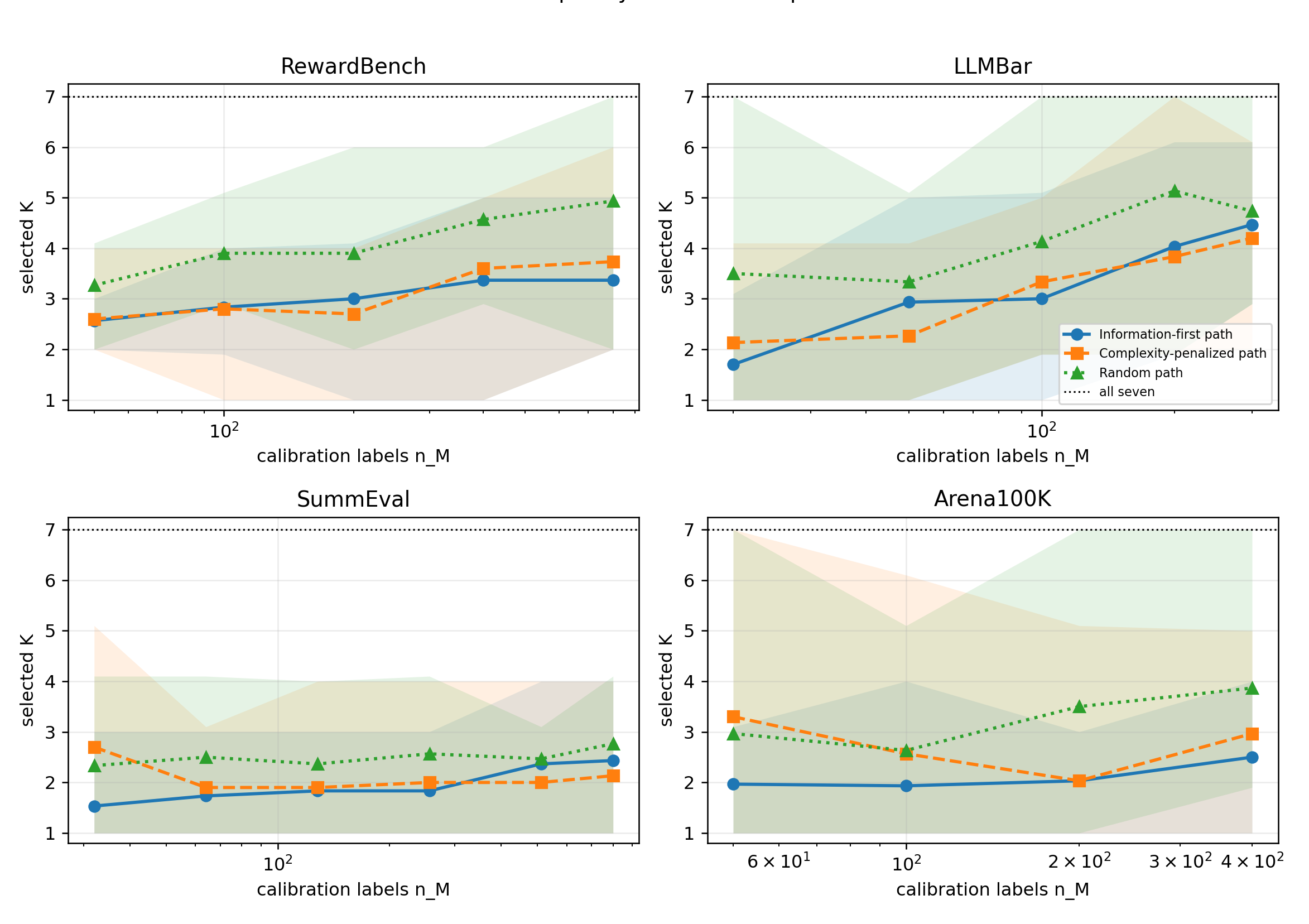}
    \caption{Selected prefix.}
  \end{subfigure}
  \caption{Uncertainty versions of Figures~\ref{fig:risk-k} and
  \ref{fig:selected-k}. Lines show split means and shaded bands show the 10--90\%
  split range, making finite-risk non-monotonicity and split-variable selected
  complexity visible.}
  \label{fig:uncertainty-diagnostics}
\end{figure}


\begin{thebibliography}{10}

\bibitem[Alam et~al.(2026)]{alam2026beyond}
Firoj Alam, Gagan Bhatia, Sahinur Rahman Laskar, and Shammur Absar Chowdhury.
\newblock Beyond LLM-as-a-judge: Deterministic metrics for multilingual generative
text evaluation.
\newblock arXiv:2604.05083, 2026.

\bibitem[Alexandru et~al.(2025)]{alexandru2025selene}
Andrei Alexandru, Antonia Calvi, Henry Broomfield, Jackson Golden, Kyle Dai,
Mathias Leys, Maurice Burger, Max Bartolo, Roman Engeler, Sashank Pisupati,
Toby Drane, and Young Sun Park.
\newblock Atla Selene Mini: A general purpose evaluation model.
\newblock arXiv:2501.17195, 2025.

\bibitem[Austin et~al.(2021)]{austin2021program}
Jacob Austin, Augustus Odena, Maxwell Nye, Maarten Bosma, Henryk Michalewski,
David Dohan, Ellen Jiang, Carrie Cai, Michael Terry, Quoc Le, and Charles Sutton.
\newblock Program synthesis with large language models.
\newblock arXiv:2108.07732, 2021.

\bibitem[Bartlett and Mendelson(2002)]{bartlett2002rademacher}
Peter~L. Bartlett and Shahar Mendelson.
\newblock Rademacher and Gaussian complexities: Risk bounds and structural results.
\newblock \emph{Journal of Machine Learning Research}, 3:463--482, 2002.

\bibitem[Khomiakov and Frellsen(2026)]{khomiakov2026noise}
Maxim Khomiakov and Jes Frellsen.
\newblock Noise-response calibration: A causal intervention protocol for LLM-judges.
\newblock arXiv:2603.17172, 2026.

\bibitem[Badshah et~al.(2026)]{badshah2026scope}
Sher Badshah, Ali Emami, and Hassan Sajjad.
\newblock SCOPE: Selective conformal optimized pairwise LLM judging.
\newblock arXiv:2602.13110, 2026.

\bibitem[Chiang et~al.(2024)]{chiang2024chatbotarena}
Wei-Lin Chiang, Lianmin Zheng, Ying Sheng, Anastasios~N. Angelopoulos, Tianle Li,
Dacheng Li, Hao Zhang, Banghua Zhu, Michael~I. Jordan, Joseph~E. Gonzalez, and
Ion Stoica.
\newblock Chatbot Arena: An open platform for evaluating LLMs by human preference.
\newblock arXiv:2403.04132, 2024.

\bibitem[Chao et~al.(2024)]{chao2024jailbreakbench}
Patrick Chao, Edoardo Debenedetti, Alexander Robey, Maksym Andriushchenko,
Francesco Croce, Vikash Sehwag, Edgar Dobriban, Nicolas Flammarion,
George~J. Pappas, Florian Tram{\`e}r, Hamed Hassani, and Eric Wong.
\newblock JailbreakBench: An open robustness benchmark for jailbreaking large
language models.
\newblock In \emph{NeurIPS Datasets and Benchmarks Track}, 2024.

\bibitem[Cobbe et~al.(2021)]{cobbe2021training}
Karl Cobbe, Vineet Kosaraju, Mohammad Bavarian, Mark Chen, Heewoo Jun,
Lukasz Kaiser, Matthias Plappert, Jerry Tworek, Jacob Hilton, Reiichiro Nakano,
Christopher Hesse, and John Schulman.
\newblock Training verifiers to solve math word problems.
\newblock arXiv:2110.14168, 2021.

\bibitem[Dawid(1982)]{dawid1982calibrated}
A.~Philip Dawid.
\newblock The well-calibrated Bayesian.
\newblock \emph{Journal of the American Statistical Association}, 77(379):605--610,
1982.

\bibitem[Dawid and Skene(1979)]{dawid1979maximum}
A.~Philip Dawid and Allan~M. Skene.
\newblock Maximum likelihood estimation of observer error-rates using the EM algorithm.
\newblock \emph{Applied Statistics}, 28(1):20--28, 1979.

\bibitem[Dadkhahi et~al.(2025)]{dadkhahi2025distribution}
Hamid Dadkhahi, Firas Trabelsi, Parker Riley, Juraj Juraska, and Mehdi
Mirzazadeh.
\newblock Distribution-calibrated inference time compute for thinking
LLM-as-a-judge.
\newblock arXiv:2512.03019, 2025.

\bibitem[DeepSeek-AI(2026)]{deepseekai2026deepseekv4}
DeepSeek-AI.
\newblock DeepSeek-V4: Towards highly efficient million-token context intelligence.
\newblock Model card and technical report, 2026. \url{https://huggingface.co/deepseek-ai/DeepSeek-V4-Flash}.

\bibitem[Dev et~al.(2026)]{dev2026judge}
Sunishchal Dev, Andrew Sloan, Joshua Kavner, Nicholas Kong, and Morgan Sandler.
\newblock Judge Reliability Harness: Stress testing the reliability of LLM judges.
\newblock arXiv:2603.05399, 2026.

\bibitem[Fabbri et~al.(2021)]{fabbri2021summeval}
Alexander~R. Fabbri, Wojciech Kryscinski, Bryan McCann, Caiming Xiong, Richard
Socher, and Dragomir Radev.
\newblock SummEval: Re-evaluating summarization evaluation.
\newblock \emph{Transactions of the Association for Computational Linguistics},
9:391--409, 2021.

\bibitem[Feng et~al.(2026)]{feng2026noisy}
Chen Feng, Minghe Shen, Ananth Balashankar, Carsten Gerner-Beuerle, and
Miguel~R.~D. Rodrigues.
\newblock Noisy but valid: Robust statistical evaluation of LLMs with imperfect
judges.
\newblock In \emph{Proceedings of ICLR}, 2026.

\bibitem[Fu et~al.(2023)]{fu2023reliable}
Xue-Yong Fu, Md~Tahmid Rahman Laskar, Cheng Chen, and Shashi Bhushan Tn.
\newblock Are large language models reliable judges? A study on the factuality
evaluation capabilities of LLMs.
\newblock In \emph{Proceedings of the Third Workshop on Natural Language Generation,
Evaluation, and Metrics (GEM)}, pages 310--316, 2023.

\bibitem[Gemma Team(2025)]{gemmateam2025gemma3}
Gemma Team.
\newblock Gemma 3 technical report.
\newblock arXiv:2503.19786, 2025.

\bibitem[Grattafiori et~al.(2024)]{grattafiori2024llama3}
Aaron Grattafiori et~al.
\newblock The Llama 3 herd of models.
\newblock arXiv:2407.21783, 2024.

\bibitem[Guo et~al.(2017)]{guo2017calibration}
Chuan Guo, Geoff Pleiss, Yu Sun, and Kilian~Q. Weinberger.
\newblock On calibration of modern neural networks.
\newblock In \emph{Proceedings of the 34th International Conference on Machine
Learning}, 2017.

\bibitem[Gy{\"o}rfi et~al.(2002)]{gyorfi2002distribution}
L{\'a}szl{\'o} Gy{\"o}rfi, Michael Kohler, Adam Krzy{\.z}ak, and Harro Walk.
\newblock \emph{A Distribution-Free Theory of Nonparametric Regression}.
\newblock Springer, 2002.

\bibitem[Hashemi et~al.(2024)]{hashemi2024llmrubric}
Helia Hashemi, Jason Eisner, Corby Rosset, Benjamin Van Durme, and Chris Kedzie.
\newblock LLM-Rubric: A multidimensional, calibrated approach to automated
evaluation of natural language texts.
\newblock In \emph{Proceedings of ACL}, 2024.

\bibitem[Hoeffding(1963)]{hoeffding1963probability}
Wassily Hoeffding.
\newblock Probability inequalities for sums of bounded random variables.
\newblock \emph{Journal of the American Statistical Association}, 58(301):13--30,
1963.

\bibitem[Jain et~al.(2025)]{jain2025beyond}
Suryaansh Jain, Umair~Z. Ahmed, Shubham Sahai, and Ben Leong.
\newblock Beyond consensus: Mitigating the agreeableness bias in LLM judge
evaluations.
\newblock arXiv:2510.11822, 2025.

\bibitem[Jiang et~al.(2023)]{jiang2023mistral}
Albert~Q. Jiang, Alexandre Sablayrolles, Arthur Mensch, Chris Bamford,
Devendra~S. Chaplot, Diego de~las Casas, Florian Bressand, Gianna Lengyel,
Guillaume Lample, Lucile Saulnier, L\'{e}lio Renard Lavaud, Marie-Anne Lachaux,
Pierre Stock, Teven Le~Scao, Thibaut Lavril, Thomas Wang, Timoth\'{e}e Lacroix,
and William El~Sayed.
\newblock Mistral 7B.
\newblock arXiv:2310.06825, 2023.

\bibitem[Kim et~al.(2024)]{kim2024prometheus2}
Seungone Kim, Juyoung Suk, Shayne Longpre, Bill Yuchen Lin, Jamin Shin,
Sean Welleck, Graham Neubig, Moontae Lee, Kyungjae Lee, and Minjoon Seo.
\newblock Prometheus 2: An open source language model specialized in evaluating
other language models.
\newblock In \emph{Proceedings of EMNLP}, 2024.

\bibitem[Kohli(2026)]{kohli2026nine}
Guneet Kohli.
\newblock Nine judges, two effective votes: Correlated errors undermine LLM
evaluation panels.
\newblock arXiv:2605.29800, 2026.

\bibitem[Lambert et~al.(2024)]{lambert2024rewardbench}
Nathan Lambert, Valentina Pyatkin, Jacob Morrison, LJ Miranda, Bill Yuchen Lin,
Khyathi Chandu, Nouha Dziri, Sachin Kumar, Tom Zick, Yejin Choi, Noah~A. Smith,
and Hannaneh Hajishirzi.
\newblock RewardBench: Evaluating reward models for language modeling.
\newblock arXiv:2403.13787, 2024.

\bibitem[Landesberg and Narayan(2025)]{landesberg2025causal}
Eddie Landesberg and Manjari Narayan.
\newblock Causal judge evaluation: Calibrated surrogate metrics for LLM systems.
\newblock arXiv:2512.11150, 2025.

\bibitem[Lail and Markham(2026)]{lail2026cost}
Ryan Lail and Luke Markham.
\newblock On cost-effective LLM-as-a-judge improvement techniques.
\newblock arXiv:2604.13717, 2026.

\bibitem[Li(2026)]{li2026calibrate}
Yanran Li.
\newblock Calibrate, don't curate: Label-efficient estimation from noisy LLM
judges.
\newblock arXiv:2605.09702, 2026.

\bibitem[Liu et~al.(2023)]{liu2023geval}
Yang Liu, Dan Iter, Yichong Xu, Shuohang Wang, Ruochen Xu, and Chenguang Zhu.
\newblock G-Eval: NLG evaluation using GPT-4 with better human alignment.
\newblock In \emph{Proceedings of EMNLP}, 2023.

\bibitem[Platt(1999)]{platt1999probabilistic}
John~C. Platt.
\newblock Probabilistic outputs for support vector machines and comparisons to
regularized likelihood methods.
\newblock In \emph{Advances in Large Margin Classifiers}, 1999.

\bibitem[Qwen Team(2024)]{qwen2024qwen25}
Qwen Team.
\newblock Qwen2.5 technical report.
\newblock arXiv:2412.15115, 2024.

\bibitem[Qian et~al.(2026)]{qian2026trust}
Mengjie Qian, Guangzhi Sun, Mark~J.~F. Gales, and Kate~M. Knill.
\newblock Who can we trust? LLM-as-a-jury for comparative assessment.
\newblock arXiv:2602.16610, 2026.

\bibitem[Radharapu et~al.(2025)]{radharapu2025calibrating}
Bhaktipriya Radharapu, Eshika Saxena, Kenneth Li, Chenxi Whitehouse, Adina
Williams, and Nicola Cancedda.
\newblock Calibrating LLM judges: Linear probes for fast and reliable uncertainty
estimation.
\newblock arXiv:2512.22245, 2025.

\bibitem[Sheng et~al.(2025)]{sheng2025analyzing}
Huanxin Sheng, Xinyi Liu, Hangfeng He, Jieyu Zhao, and Jian Kang.
\newblock Analyzing uncertainty of LLM-as-a-judge: Interval evaluations with
conformal prediction.
\newblock In \emph{Proceedings of EMNLP}, 2025.

\bibitem[Stone(1974)]{stone1974cross}
Mervyn Stone.
\newblock Cross-validatory choice and assessment of statistical predictions.
\newblock \emph{Journal of the Royal Statistical Society: Series B}, 36(2):111--133,
1974.

\bibitem[Tang et~al.(2025)]{tang2025arenaexplorer}
Kelly Tang, Wei-Lin Chiang, and Anastasios~N. Angelopoulos.
\newblock Arena Explorer: A topic modeling pipeline for LLM evals and analytics.
\newblock Dataset citation for \texttt{lmarena-ai/arena-human-preference-100k}, 2025.

\bibitem[Wolpert(1992)]{wolpert1992stacked}
David~H. Wolpert.
\newblock Stacked generalization.
\newblock \emph{Neural Networks}, 5(2):241--259, 1992.

\bibitem[Yu et~al.(2026)]{yu2026heterogeneous}
Shibo Yu, Yingzhou Wang, Yan Chen, Guodong Li, and Jin-Hong Du.
\newblock Heterogeneous judge-aware ranking with sensitivity, disagreement, and
confidence.
\newblock arXiv:2605.05073, 2026.

\bibitem[Zeng et~al.(2024)]{zeng2024llmbar}
Zhiyuan Zeng, Jiatong Yu, Tianyu Gao, Yu Meng, Tanya Goyal, and Danqi Chen.
\newblock Evaluating large language models at evaluating instruction following.
\newblock In \emph{Proceedings of ICLR}, 2024.

\bibitem[Verga et~al.(2024)]{verga2024juries}
Pat Verga, Sebastian Hofst{\"a}tter, Sophia Althammer, Yixuan Su, Aleksandra
Piktus, Arkady Arkhangorodsky, Minjie Xu, Naomi White, and Patrick Lewis.
\newblock Replacing judges with juries: Evaluating LLM generations with a panel
of diverse models.
\newblock arXiv:2404.18796, 2024.

\bibitem[Zhu et~al.(2023)]{zhu2023judgelm}
Lianghui Zhu, Xinggang Wang, and Xinlong Wang.
\newblock JudgeLM: Fine-tuned large language models are scalable judges.
\newblock arXiv:2310.17631, 2023.

\bibitem[Zheng et~al.(2023)]{zheng2023judging}
Lianmin Zheng, Wei-Lin Chiang, Ying Sheng, Siyuan Zhuang, Zhanghao Wu, Yonghao
Zhuang, Zi Lin, Zhuohan Li, Dacheng Li, Eric~P. Xing, Hao Zhang, Joseph~E.
Gonzalez, and Ion Stoica.
\newblock Judging LLM-as-a-judge with MT-Bench and Chatbot Arena.
\newblock In \emph{Advances in Neural Information Processing Systems}, 2023.

\end{thebibliography}
\end{document}